\magnification 1200 \input epsf
\hsize 6.5truein
\vsize 9truein
\def\timeofday{\count11=\time  \count12=\count11
 \divide\count12 by 60 \count14=\count11   \count13=\count12
 \multiply\count13 by 60 \advance\count14 by -\count13}
\footline={\rm{\the\month\slash\the\day\slash\the\year\hskip .2in
 \the\count12 :\ifnum\count14<10{0}\fi\the\count14\hfill\folio}}
\baselineskip=12pt
\parskip=3truept
\hfuzz=10pt
\timeofday
\vbox to \vsize{\vfill
\centerline{\bf Reworking geometric morphometrics}

\centerline{\bf into a methodology of transformation grids}
\vskip .2in
\centerline{Fred L. Bookstein}

\centerline{University of Vienna, University of Washington}
\vskip .4in
\noindent ORCID: 0000-0003-2716-8471
\vskip .4in
\noindent {\tt fred.bookstein@univie.ac.at, flbookst@uw.edu}
\vskip 1in

\itemitem{}This is an update of a manuscript of the same name and
 authorship that was submitted to
 {\sl Evolutionary Biology} on January 2, 2023.  

\vfill}\eject

{\bf Abstract.}
      Today's typical application of geometric morphometrics to
 a quantitative comparison of organismal anatomies begins by
 standardizing samples of homologously labelled
 point configurations for location, orientation, and scale, and then 
 renders the ensuing comparisons graphically by thin-plate spline
 as applied to group averages, principal components,
 regression predictions, or canonical variates.  The scale-standardization
 step has recently come under criticism as unnecessary and
 indeed inappropriate, at least for growth studies.  This essay
 argues for a similar rethinking of the centering and rotation, and
 then the replacement of the thin-plate spline interpolant of the
 resulting configurations by a different strategy
 that leaves unexplained residuals at every landmark individually in
 order to simplify the interpretation of the displayed grid as a whole,
 the ``transformation grid'' that has been highlighted as the true
 underlying topic ever since D'Arcy Thompson's celebrated exposition of 1917.
 For analyses of comparisons involving gradients at large
 geometric scale, this paper argues for
 replacement of all three of the Procrustes conventions by a version
 of my two-point registration of 1986 (originally
 Francis Galton's of 1907).  The choice of the two points interacts
 with another non-Procrustes concern, interpretability of the
 grid lines of a coordinate system deformed according to a fitted
 polynomial trend rather than an interpolating thin-plate spline.
 The paper works two examples using previously published midsagittal cranial
 data; there result new findings pertinent to the interpretation of
 both of these classic data sets.
 A concluding discussion suggests that the current
 toolkit of geometric morphometrics, centered on Procrustes shape
 coordinates and thin-plate splines, is too restricted to suit many of the
 interpretive purposes of evolutionary and developmental biology.
\vskip .2in

\vfill\eject
\centerline{\bf I.  Introduction}
\vskip .2in

       Figure 1 here arose simply as free play with the
 tools of geometric morphometrics (GMM).  The data set comprises the
 familiar ``Vilmann octagons'' tracing around the midsagittal
 neurocrania of close-bred laboratory rats
 radiographed in the 1960's by the Danish anatomist
 Henning Vilmann at eight
 ages between 7 days and 150 days and digitized some years later by
 the New York craniofacial biologist Melvin Moss.  
 This version of the data is the one explored in
 my textbook of 2018: the subset of 18 animals with
 complete data (all eight landmarks) at all eight ages.
 The concern of Figure 1 is the contrast of the Procrustes-averaged
 shapes for the age-7 and age-150
 animals (only the averages, no consideration of
 covariances). The heavy lines are for the age-150 data subset,
 the light lines, the data from the animals at age 7 days (a configuration
 this paper will occasionally refer to as the ``template'').
 All panels of the figure complicate the usual Procrustes plot of
 shape coordinate pairs by all or some of the segments connecting
 these coordinate pairs.  In the figure's left column, 
 all $8\cdot7/2=28$ of the interlandmark segments have been drawn;
 in  the right column, only the subset that are the reason for calling
 your attention to this figure.  In the top row, those average locations
 correspond to the usual Procrustes-registered 
 shape coordinates, partialling
 out only centering, size, and rotation. Panel (b)
 is limited just to the nine (out of 28) interlandmark
 segments from panel (a)
 that rotated either way by at least 8.6 degrees (the figure label
 expresses this as ``0.15 radians'')
\footnote{$^1$}{One
 radian is the mathematician's
 natural metric of angle, the angle (about $57^\circ$) at which the extent of a
 circular arc is equal to the radius of the circle.}
 between age 7 and 150 days. A pretty graphic, but it
 features too much overlay of signals to qualify as a legible pattern
 analysis.

 However, most of the
 clutter is due to the substantial change of aspect ratio
 (height-to-width ratio, 
 obvious in the left column) that rotated both of the longer diagonals 
 (Basion to Bregma, Lambda to SES) of the template.
       Fortunately, we already
 know how to remove this unwanted uniformity of relative
 vertical compression from our comparison: recourse to the ``nonuniform''
 component of Procrustes shape space, complement to the subspace of uniform
 transformations (those that take all rectangles into
 parallelograms).
 The resulting plots are the pair in the
 bottom row.

 It is no surprise that
 the diagram at lower left, panel (c), looks even more cluttered
 than panel (a), because now
 the calvarial roof, not just the cranial base, overlaps between the
 ages. But there is also a new signal once
 the diagram is edited to suppress all
 the segments that didn't rotate much, a signal
 that seems not to have been
 anticipated in previously published analyses of these data.
 As panel (d) shows, six of the 28 possible segments
 rotate by more than 0.15 radian 
 after standardizing this uniform 
 aspect of the young-to-old comparison.
 And now the pattern is obvious. The five landmarks at left
 (anatomical posterior, SOS around to Lam)
 are rotating clockwise (in this projection) over growth, while the
 three at the right, located anatomically anteriorly,
 are rotating counterclockwise, all this
 to a longitudinal arrangement (think of the centroid of the set of five, versus
 the centroid of the frontmost three) that isn't rotating either way.
 This paper will refer to the segmented polygon
 SOS-Bas-Opi-IPS-Lam, the set of five landmarks at left in
 Figure 1(d), as the ``posterior pentagon''
 and the remaining three, Brg-SES-ISS, as the ``anterior triangle.''

      The opposition of rotations in panel (d) is consistent with
 a report using an alternative arithmetic of
 intersegment length-ratios.  There is evidently
 shortening of upper calvarial anteroposterior length, Lam to Brg,
 relative to the central segment of the cranial base from ISS to SOS.
 Now there is no need to state that this pattern is
 ``relative to the sequestering of the uniform term,''
 as uniform transformations do not alter ratios of distances in the
 same direction, whether concurrent or parallel.
  
      This relative rotation, including that contrast of vertically aligned
 horizontal growth rates, the central cranial base
 versus the calvarial roof above it,
 is surely a feature of the 143-day change of
 form here.  But where is it to be found in the GMM toolkit?  Figure 2
 recovers exactly the same report from a quantitative style dating
 back more than 80 years prior to GMM, analysis via the coordinates
 Francis Galton introduced in 1907 for ``classification of
 portraits.''\footnote{$^2$}{The GMM literature usually refers to these as
 ``two-point coordinates'' or an ``edge registration,'' while the
 statistical literature (Stuart and Ord 1994:279) calls them 
 ``Bookstein coordinates'' in keeping with Stigler's Law, which
 states that usually innovations are named after the second person
 to stumble across them. Ignoring the scaling aspect of this tool,
 the centering and orientation here was already explicit in Boas (1905)
 and probably can be traced back all the way to the German anthropologists'
 adoption of the celebrated ``Frankfurt Horizontal'' in 1882 (see
 Garson, 1885 [which, remarkably enough, is available from JSTOR] ---
 Orbital set to $(0,0)$, Porion along the positive $x$-axis:
 the {\sl Ohr-Augen Horizontale} of Martin 1914). For a contemporary
 critique of this specific convention of 1882, see Bookstein, 2016.}
 Here I have diagrammed every possible
 two-point registration  of these octagons (quantified only by
 their average coordinates as Moss originally digitized them).
 For each alternative baseline, the original
 Cartesian coordinate average configuration has been separately
 rotated and scaled so that the first baseline point is at $(0,0)$ of
 a new coordinate system and the second is at $(1,0)$ in the
 same system (the two points circled in every panel of the
 figure).  We have thereby altered every single step of the
 Procrustes toolkit --- the centering, the rotating, the scaling --- while eschewing
 any recourse to the thin-plate spline for separating out that uniform
 term.
 And yet ten of the panels clearly show the same phenomenon, the
 relative rotation between the anatomically posterior pentagon of landmarks
 and the anterior triangle.  
 Whenever both ends of the baseline are in the same sector (here numbered
 [8,1,2,3,4] versus [5,6,7]), the
 rotation is clear in the behavior of the complementary sector.
 This is particularly evident in the analysis to baselines 5-6 (row 4 column 5),
 5-7 (row 4 column 6), or 3-4 (row 3 column 2), where, regardless of any
 overall change of aspect ratio, the border
 of the octagon opposite the baseline appears
 to have radically shifted by a rotation with respect to that baseline.
 The disparity  between ratios of change of length for segments ISS-SOS and
 Lam-Brg is clearest, perhaps, in the panel for that ISS-SOS baseline,
 fifth row, fourth column.

       Such an analysis, both elegant and elementary, shares no
 arithmetic with the standard GMM toolkit of Procrustes registration and
 thin-plate splines. 
 (For a good overview of computational aspects of
 that standard toolkit in a format suitable for routine
 biometric applications, see Claude 2008.) 
 It is far older than that morphometric synthesis
 of the 1990's, older even than analysis by triangles (``tensor
 biometrics,'' Bookstein, 1984) or
 by biorthogonal grids (Bookstein, 1978). Both of these versions involve
 attention to short or long transects
 of the form that intersect {\sl internally,}
 where, by analogy with the change of form from a square to
 a rectangle, for one particular pair of
 directions (sides of the square)
 the ratios of change of distance are greatest or least and
 the angle of intersection is
 invariant at  $90^\circ,$ while the ratio of change of
 the two distances at $45^\circ$ to these directions 
 (diagonals of the square) is unity and it is
 the change of their {\sl angle} that is maximized.  A closer inspection
 of the interlandmark-distance interpretation of Figure 1(d) instead
 makes reference to distances that are parallel at some spacing (upper calvarial
 width versus lower), a change visible equally in the Procrustes fits and
 in the two-point versions, especially versions 7-8 (row 5 column 4)
 and 3-5 (row 3 column 3).  The idea of examining ratios of parallel distances
 like these is already present in some much earlier applied treatises, such as
 Martin 1914.  

       For an intuitive understanding of what is going on here, turn back
 to the earliest textbook introduction of the thin-plate spline, Bookstein 1991, where analyses
 like these, restricted to just a quadrilateral of landmarks, exemplify
 what I called ``purely inhomogeneous transformations'' there,
 meaning, transformations without any uniform component.
 Figure 7.3.6 of that book displays, within the limits of
 the software tools of the time, the effect of rotating the
 starting grid on the graphs of this purely inhomogeneous component
 (here, the sole nonlinear component)
 of the deformations of a square that
 minimize net bending energy --- the now-ubiquitous thin-plate spline.
 
Figure 3 is a modification of that textbook figure intended to clarify
 the contrast of the different types of salience
 (length-ratios and rotations)
 for the pairs of segments of interest.
 Its four columns prototype different types of the transformations, each
 of a starting square of landmarks.
 In the top row are the starting squares, twice in Cartesian alignment with
 the page and twice at $45^\circ$.  Below are the corresponding analyses,
 enhanced by ordinary thin-plate splines that are not actually part of
 the arithmetical report.
  (But that spline has nothing to do with the analysis
 here, which deals only with the landmark positions per se, not
 any interstitial tissue. The quadratic extension to the interstitial rendering
 in Figures 4 through 11 requires
 a minimum of six landmarks, not just these four; the further extension to
 a cubic fit in Figures 17 and 19 requires at least ten.) 
  In column (a) the square is transformed to
 a rhombus by rotating two of its edges without change in length.
 What change are the angles between the concurrent edges.  In column (b)
 the same transformation is applied to the square of landmarks at
 $45^\circ$ (in other words, the grid has rotated with respect to
 the landmarks, the configuration of which has not changed
 in either row).  Now the report is reversed: the greatest change is in
 the ratio of lengths of diagonals, while the angle between them is
 left invariant at $90^\circ.$ This was also the case for the configuration
 in column 1, where it was confounded by the inconvenient orientation of
 the grid lines. 

The situation in Figure 1(d) or Figure 2 corresponds instead to the prototype
 in columns (c) or (d) of Figure 3. 
 The starting configuration is still the same
 square.  But in column (c) the transformation changes the ratio of 
 lengths of two edges that are {\sl parallel} (horizontal in the figure),
 not perpendicular as
 in columns (a) or (b), while leaving unchanged the ratio of the other
 two edge lengths (the other pair of
 parallels in panel c1) while radically altering their
 {\sl angle.}  This is a transformation from a square to an isosceles
 trapezoid. The complementary transformation in column (d), which the
 geometer would call square-to-kite, leaves
 the diagonals unchanged in length and in angle while altering the
 relation between their midpoints.  Now it is a different pair of paired
 edges whose length-ratio has not changed --- the top and bottom $V$'s ---
 and while the angles at the end of the horizontal diagonal are hardly altered,
 those at the ends of the vertical diagonal are greatly changed, one
 increased and the other decreased.  To repeat, these reports rely not
 at all on any GMM technology, neither Procrustes nor thin-plate spline.

The aim of this paper is to push this insight as far as it can
 go while remaining elementary in its biomathematics. (For instance,
 its multivariate analysis is limited to the familiar setting
 of multiple regression.)  While the idea
 of two-point coordinates was originally Galton's in 1907, the idea
 at which the analysis here is aimed, the quadratic growth-gradient,
 is only half as old: it is present in embryo in Peter Sneath's underrated paper of 1967 on
 trend-surface analysis of D'Arcy Thompson's transformation grids. 
 The core of the argument inheres in any of the next eight figures, which
 selected eight interesting baselines from the 28 in Figure 2 for
 expansion of the analysis to include an explicit quadratic regression
 of the averaged age-150 Cartesian coordinates against the same
 from the age-7 octagons.  These analyses completely ignore the tools
 of standard GMM --- there is no Procrustes centering, no scaling  or
 reorientation beyond the (arbitrary) choice of baseline, and 
 thin-plate splines are drawn only to be dismissed --- while
 what results, you will see, is a coherent summary of this particular
 change of neurocranial form. A combined Figure 12 
 arrays the eight separate summaries for a synthesis of
 their information  content abstracted in Figures 13 and 14.
  Following this exploration,
 a further analysis of some data from a study of cranial hominization
 my Vienna group published twenty years ago will consider some
 extensions of this approach, and a concluding Discussion will reflect
 on some implications of this seeming irrelevance of today's
 conventional GMM toolkit
 for the explanatory purposes of evolutionary or developmental morphology.

\vskip .2in
\centerline{\bf II. Vilmann 7-to-150-day growth analyzed without Procrustes GMM}
\vskip .2in

      The recommended alternate analysis of the Vilmann growth gradient
 in Figure 1 may be narrated by an extraction
 of common findings from a suite of separate analyses to my selection of 
 baselines, some transects of the octagon and others circumferential to it.
 The analyses to be synthesized are laid out in Figures 4 through 11.

      Each of these eight composites offers four panels.
 At upper left will be the conventional thin-plate spline of
 the averaged octagon of Cartesian coordinates of the age-7-day animals
 as warped into the analogous average at age 150 days. Analysis is
 to the baseline of the pair of landmarks indicated in larger circles,
 which specifies the orientation of the thin-plate spline grid in
 each example.  To its right will be a gridded version (in this orientation)
 of the ``growth fit'' likewise displayed first for comparison
 as a thin-plate spline. Here the $x$-coordinate of the deformed
 grid is the predicted $x$-coordinate
 from the regression of the baseline-standardized octagon vertices of
 the 150-day average on {\sl both} coordinates of the age-7 average,
 and also their squares and their crossproduct (i.e., a regression
 of each $x_{150}$ on $(x_7,y_7,x_7^{~2},y_7^{~2},x_7y_7)$) 
 and likewise the $y_{150}$-coordinates. Each of these regressions
 involves five predictors, plus a constant, for only eight ``cases''
 (the relevant coordinate, $x$ or $y$, of the eight landmarks),
 and so has only two degrees of freedom for error; they are not
 really regressions, but rather almost interpolations, when the landmark count
 is so small. At left in the lower
 row will be a more appropriate representation of the quadratic fit splined
 at above right: actual transforms of the grid lines of the starting form to
 this baseline, with the regressions' ``dependent variable'' the $x-$ and
 $y-$coordinates of the filled dots corresponding to the fitted locations at
 the open circles nearby.  Finally, the panel at lower right will restrict this 
 grid to just the interior of the age-7 octagon --- this portion of the
 graphic deserves attention first, before any extensions to the exterior.

 Consider, then, the first figure in this series, Figure 4, which is
 the analysis for a baseline from Basion to Opisthion --- the shortest
 interlandmark segment in the template, but one contained entirely
 within the posterior pentagonal compartment of Figure 1(d) and hence
 one that might enlighten us as to the rotation archived there.
 That rotation between anterior landmark triangle and posterior landmark
 pentagon is essentially the same that
 is displayed in Figure 1 for the conventional GMM approach and in
 Figure 2 for the panel corresponding to this baseline (there, the panel
 in row 1, column 1) --- indeed it will be the same in all eight
 of the series of figures. Here in Figure 4, for the baseline Basion to Opisthion
 (axis of the midsagittal foramen magnum), the orientation of the form is
 rotated about $130^\circ$ from the Procrustes convention in Figure 1.
 The thin-plate spline (upper left panel)
 is interesting in that inside the posterior
 five-landmark component, SOS around to Lam, the interior as rendered
 by the spline appears to be nearly affine (all grid cells the same
 size and shape) except near IPS,
 and likewise nearly affine
 for the anterior component Bas-SES-ISS.  The growth
 fit (upper right panel) apparently has pulled IPS to the left in
 this diagram. As Figure 1 shows, and as has been exposited in earlier
 papers (e.g., Bookstein 2017), this
 point participates in a specific focal process displacing it upward
 in the more realistic anatomical setting of Figure 1.  Thus the fit in 
 this upper right panel of Figure 4 does not show the deviation of change at
 IPS from change at its neighbors that is present in the actual data.   

Either panel of the lower row shows how closely
 the fitted landmarks (open circles)
 track the averaged 150-day locations observed (the solid circles).
 The horizontal grid lines in the interior
 of the form (lower right panel) are mainly
 straight, while their orientation  on this diagram is graded from top
 to bottom more smoothly than one would infer from the analogous diagram
 at upper left (the thin-plate spline based on the fully detailed
 data record, which, by design, is not conducive to any lower-dimensional
 summary).  The steady rotation of this imputed grid line direction is complemented
 by a gentle curvature of the {\sl other} grid line direction, a curvature that
 is not so apparent in the explicit thin-plate spline at upper left ---
 the transformation that appears segmented there, one nearly linear system 
 for the posterior pentagon and another for the anterior triangle,
 is smoothed by the quadratic regression into a continuous gradient
 from end to end of the template (top to bottom of the grid, in
 this coordinate system).
 Note that the prediction error of the quadratic fit (lower row)
 specifically implicates the length of the chosen baseline,
 at both ends.

Figure 5 shows the same analysis for a different baseline, Basion to 
 Interparietal suture, from the same posterior pentagon.  
 Again the quadratic fit (lower row) shows a substantial residual,
 this time at only one of the baseline points (IPS).  In the deformed
 grid, both systems of lines are curved, a feature that makes
 interpretation more difficult.

Figure 6 is the first to involve a cross-component baseline, Basion
 (from the posterior pentagon) to Bregma (from  the anterior triangle).
 The starting grid has rotated about $80^\circ$ from its position in the
 first of this series (i.e., the angle between segments Basion-Opisthion
 and Basion-Bregma in the age-7 average is about $80^\circ$).
 Again the panels in the lower row inform us that the initially
 vertical grid lines (lines along Lambda-ISS or IPS-SOS)
 are transformed by the quadratic fit into a
 pencil of nearly straight lines at varying orientations,
 while the lines of the originally orthogonal system are gently curved 
 in a manner that will concern us in detail in Figure 13.
  At neither of the baseline
 points is there any substantial fitting error of the quadratic
 regressions.
 The approximate uniformity of cell sizes across the
 trimmed grid at lower right here and in every other figure of
 this series assures us that the recourse to distances from the centroid in
 models of centric allometry, such as Bookstein 2021a, is a reasonable
 default.   Indeed the separation between the actual age-150 centroid
 and the quadratic trend transform of the age-7 centroid is a mere
 0.054 units in the scale of this figure.

Figure 7, to a baseline from IPS to SOS, is very nearly the same
 analysis as in Figure 6 inasmuch as the two baselines, IPS-SOS
 and Bas-Brg, are nearly at $90^\circ$ in the age-7 template.
 The main difference is the substantial increase in fitting error,
 owing to the fact that landmark 3, IPS, is known to be strongly loaded on a
 special factor not shared with the rest of the configuration.  
 Nevertheless, the grids of the lower row still greatly resemble
 those of the preceding figure, for the baseline at $90^\circ$ to this one:
 lines parallel to IPS-ISS (here, the baseline) remain straight
 but rotate from end to end, while the orthogonals are gently
 curved.

Let us move more quickly through the remaining versions of this
 four-panel scheme.  In Figure 8, baseline Lambda-SOS, both
 systems of grid lines are gently curved (although the rotation
 from end to end of the original octagon is as clear as if they
 had remained straight).  The errors of fit at the baseline points
 are moderate in magnitude, partly because the fit at Lambda
 is distorted by the need to accommodate the deviation at
 IPS.
 
Figure 9, for a baseline Bregma-SES within the {\sl anterior} component
 of Figure 1, displays gentle curves in both grid systems.
 Errors of the quadratic fit are again moderate, and the rotation
 so evident in Figure 1(d) is very clear
 in spite of the curvature of these deformed grid lines.  
 The baseline in Figure 10, Bregma-ISS, has similar errors of
 fit and similar curving of the grid lines.  Finally, Figure 11,
 for an ISS-Bas baseline, is roughtly the $90^\circ$ rotation of
 the analysis in Figure 5, whose baseline (Bas-IPS) is roughly
 at $90^\circ$ to the baseline here.  

Figure 12 summarizes all eight of these analyses in a way that
 permits some criteria of interpretability to emerge regarding
 replacement of the Procrustes
 rotation by a protocol more conducive to reportage: a protocol
 that associates the reorientation  of specimens to the
 ultimate simplification of their deformation by reference
 to the specific coordinate lines as deformed from the template's square grid.
 We have seen that baseline analyses can sometime come in pairs if the
 corresponding interlandmark segments themselves lie 
 at approximately $90^\circ,$ and it is better if they run close to
 the centroid of the octagon.
 More subtly, morphological comparisons
 that can result in reports of relative rotations
 of parts of a landmark configuration may be diagrammed best not by a 
 thin-plate spline but by a choice of a specific baseline that
 highlights the rotation in question, like Figure 6 or Figure 7 here,
 by leaving one set of grid lines straight lines even as they
 are rotated.  (For instance, in Figure 6, the panel at lower right is more
 interpretable than the panel at upper left, even though the
 information content is effectively the same.)
 The thin-plate renderings in Figure 12, columns 1 and 3, all confirm the
 relative rotation detectable already in Figure 1, but do not otherwise
 appear to offer much intuitive accessibility.  By comparison, the
 quadratic-fit displays, columns 2 and 4, vary enough in their
 legibility that some are truly insightful. 
 Those that seem most helpful are the pair of analyses
 in the second row, to baseline Bas-Lam or IPS-SOS (two directions
 that happen to be nearly perpendicular) --- these seem to be
 considerably better than the standard GMM analysis 
 at showing a potentially meaningful gradient for the
 growth process being visualized here. 

The analysis in Figure 7 suggested
 a scenario I have highlighted  in Figure 13 by the
 simple trick of extending the domain of the quadratic fit beyond
 the bounds of the landmark locations being fitted.
 The diagram here extends the earlier gridded
 transformation merely by evaluating it on the new real estate to the left in the
 same template coordinate system, i.e. into the empty space some distance above
 the foramen magnum of these animals, where the horizontal grid lines of Figure 7
 appear to be converging.  We see that the near-linearity
 of the transformation along the baseline and all the grid lines
 parallel to that direction persists quite far beyond the actual
 anatomical limits of the comparison, resulting in the strong impression
 of some sort of descriptive center at an unphysiological
 distance outside the actual calva.  The apparent rotation suggested
 in Figure 1(d) is embedded here in a larger system of reorientations
 that might be viewed as continuous rather than segmented, or,
 in a more suggestive language, graded rather than modular. The
 suggestion is strong, for instance, that this grading ought to be checked for
 extending further anteriorly to the facial skeleton (a description that
 will be tied to the classic interpretation of
 ``orthocephalization'' by a footnote in Section IV) or other features
 outside of this particular neurocranial data set. 

Figure 14 sketches two geometrical interpretations of Figure 13, one
 more familiar to the applied mathematician and the other less so.
 Each is an alternative to the thin-plate spline of column (d)
 in Figure 3; one will prove more realistic than the other for
 this paper's examples.
 The more familiar map is the {\sl projection,} central panel, that
 takes every straight line onto another straight line. But this
 mapping substantially alters the spacing of the points where these
 deformed grid lines meet the bounding kite.
 No such respacing appears
 in the extended quadratic fit itself, Figure 13. An alternative
 better matching that observed quadratic fit is the family prototyped
 in the right-hand panel of the figure, the {\sl bilinear map}a
 that I discussed in considerable detail in Bookstein 1985.
 Bilinear maps\footnote{$^3$}{In finite-element analysis, these
 are often called {\sl isoparametric coordinates} of the 
 quadrilateral.} take one quadrilateral
 onto another as follows. Every point $(x,y)$ in the interior of the template
 quadrilateral is the intersection of two lines connecting
 opposite edges that divide
 those edges in the same ratio. The map takes $(x,y)$ to the
 intersection of the two lines that divide the homologous 
 pair of edges in the target in the same ratio. The bilinear map
 of square onto kite can be written $(x,y)\rightarrow (x,y)+a(1+xy,1+xy)$
 for some $a$.  The projection in Figure 14
 required the upper isosceles triangle of
 the template to be mapped into the space above the horizontal diagonal
 of the kite, entailing a considerable compression of its vertical
 coordinate; the bilinear transformation enforces much less compression
 here, at the cost of bending that horizontal diagonal over the
 course of the deformation. This attenuation of the variability of
 those ratios of area change seems to match the graphics of all the
 quadratic fits in Figures 4 through 11 after an appropriate rotation.  

Returning one final time
 to the scheme in Figure 1(d), the decomposition of the
 neurocranial octagon into two nonoverlapping components,
 we see that the figure has indeed oversimplified the
 situation there.  That the rotation of all edges of the
 posterior pentagon leaps to the viewer's eye obscures the fact that
 all but one of these segments have changed their {\sl length.}
 And likewise the anterior ``triangle,'' Brg-SES-ISS, does
 not rotate rigidly --- its edge from SES to ISS shortens and also
 does not rotate as far as the other two.  Any report focusing
 on the two ``components'' is deficient in failing to refer to
 the coordinate space in-between them, where the unconformity
 between anteroposterior changes of length along the cranial
 base versus along the calvarial roof seems better captured by
 the rotating lines of the Bas-Brg baseline and IPS-SOS baseline
 analyses (Figure 12, row 2, columns 2 and 4) 
 than by the irregularities of the corresponding thin-plate
 splines (columns 1 and 3). 

\vskip .2in
\centerline{\bf III.  An example from hominization of the skull}
\vskip .2in
      The Vilmann analysis of Section II exploited the best study design that
 experimental zoomorphology has to offer: a sample of close-bred animals
 imaged by identical machinery at a fixed sequence of
 developmental ages. (The identification of this research
 design as the {\sl summum bonum} of laboratory evo-devo research is a
 century old --- it dates from no later than Przibram 1922.)
 Most of the data structures to which GMM has been applied are not so
 elegantly designed. This paper's final example is 
 a pair of comparisons, each much more typical in its
 design, that share one 20-landmark configuration scheme. 
 The data are a selection from the 29 forms analyzed in
 Chapter 4 of Weber and Bookstein (2011) that
 originated in computed midsagittal sections of
 a larger sample of CT scans digitized by
 Philipp Gunz for the growth analysis in Bookstein et al. (2003).
 That original analysis explicitly relied upon the
 same GMM toolkit that is most commonly invoked today: Procrustes analysis,
 principal components of the resulting shape coordinates, and visualizations
 by thin-plate spline.\footnote{$^4$}{In one version or another these data
 have already been used for demonstrations of GMM
 in textbooks three different times: 
 not only Weber and Bookstein (2011) but also Bookstein (2014, 2018).
 The approach circumventing those typical GMM maneuvers is new to the
 present paper.}  

       Of the specimens homologously digitized in 2003,
 most are {\sl Homo sapiens}, while four are named specimens of 
 {\sl H. neanderthalensis} (Atapuerca, Kabwe, Guattari, and Petralona),
 and two are specimens of {\sl Pan,}
 one of each sex.  For the present reanalysis 
 I have averaged the 18 adult {\sl sapiens} (one of which, Mlade\v c,
 is an archaic specimen) and, as a separate group,
 the four neanderthals.  As a third ``group'' (present for a didactic
 purpose, a comparison of comparisons)
 I selected the female adult chimpanzee, because the adult male shows even more
 of the heterochrony that will render my final figure so extreme
 in certain aspects of its geometry. Of course these samples are far
 more limited than any data resources that would be brought to bear
 on the same comparisons today.
 It would be unreasonable to claim that the computations to be reported
 presently are valid empirical findings; my purpose is instead to
 demonstrate a methodological alternative to Procrustes-
 and spline-based GMM.
 
The left panel of Figure 15 names these twenty landmarks at their
 positions in the average of the {\sl H. sapiens} sample in the original
 CT coordinates, which were not far from a Sella-Nasion
 orientation.  In the right panel this configuration is
 supplemented by the configurations of
 the same twenty points for
 the female chimpanzee and also for the neanderthal average, all
 after the two-point transformation (Bookstein coordinates)
 that put all three ANS's (of which two are group averages)
 at $(0,0)$ and all three internal Lambda's at $(1,0)$.  Evidently this
 coordinate system has been rotated, translated, and scaled from the
 panel at its left, but none of these steps proceeded by the Procrustes method.

    Consider first the analysis in Figure 16, which in its design echoes
 three of the four panels of the Vilmann series, Figures 4 through 11, but
 in this case only for one selected baseline, from ANS to LaI, as
 in the right panel of Figure 15.
 (Analysis to a roughly perpendicular baseline,
 Opi--BrI, results in essentially the same diagrams.) The comparison 
 in Figure 16
 is from the averaged points for {\sl H. sapiens}
 in Figure 15 to the averaged points for {\sl neanderthalensis.}
 In both of these {\sl Homo} averages (and
 also in the single female adult {\sl Pan} specimen to come)
 the baseline crosses the cranial base near Sella roughly
 halfway along its length.  The thin-plate spline deformation from the
 average of the eighteen humans to the average of the four neanderthals,
 upper left in the figure, shows the expected contrast of shrinking
 neurocranium and expanding splanchnocranium, particularly along the
 palate; the cranial base interposes itself as the so-called ``hafting zone.''
 As the upper-right panel shows, this grid is tracked to
 some extent by the analogous grid for the
 fitted values of the same {\sl neanderthalis} landmarks from the
 quadratic regression on the {\sl sapiens} coordinates, 
 That quadratic regression, already demonstrated many times
 in the Vilmann example preceding, shows most
 of its failure of fit (discrepancies between the open circles and their
 filled neighbors in  the lower-left panel) along that central separatrix,
 with a possible exception at lower right where the pairings of the two inions
 are rearranged in both separation and orientation. As Figure 15 hinted,
 this rearrangement is due mainly to excessive variation at InE, external inion.

 The final quadratic trend grid, at lower left in Figure 16, is strikingly
 different from the thin-plate spline of the same point loci (upper right).
 Indeed this grid for the fit looks remarkably
 like a rotation  of the grid at
 right in Figure 14, the bilinear transformation
 leaving two specific families of straight lines straight after
 the deformation, while their orientations rotate across.
 the diagram.  At this large scale, the
 comparison of midsagittal crania of these sister species is largely
 smooth --- the points in the hafting zone differ hardly at all from
 their predicted locations under the quadratic analysis. 
 In particular, the implication of modularity in the upper right
 panel is completely effaced in the actual quadratic fit grid at 
 lower left, indicating instead an approximating spatial process
 that is homogeneously graded with no natural
 boundaries embryological or otherwise. The grading is consistent
 with the observation that relative to the face
 the neanderthal neurocranium is
 smaller than that of {\sl sapiens} with some relative 
 rotation as well. 

Figure 17 analyzes the same comparison by a cubic fit instead
 of the quadratic fit in Figure 16.  (Specifically,
 this fit models each of the twenty $x-$coordinates of the
 {\sl H. neanderthalensis} average and then each of its twenty
 $y-$coordinates as a linear combination of {\sl nine} terms
 $x_{sap},$ $y_{sap},$ $x_{sap}^2,$ $y_{sap}^2,$ $x_{sap}y_{sap},$ 
 $x_{sap}^3,$ $ y_{sap}^3,$ $ x_{sap}^2y_{sap},$ and $x_{sap}y_{sap}^2.$
 The quadratic regressions used only the first five of these predictors.)
 These cubic grids show bizarre behavior outside the limits of their
 driving data (the strange cusps already clear in Sneath's 
 examples of 1967), so as in the Vilmann exposition of Section  II
 I extended the figure by one more panel, lower right,
 that trims the grid to just the interior of
 the region occupied by the actual target configuration
 (here the {\sl H. neanderthalensis} average).  The straight lines of
 the rendering in Figure 16 now appear as S-curves 
 across that same hafting zone, and of course the new fit, a regression on
 nine predictors, has to be closer than that in Figure 16 based on only five
 of the nine.
 But the change of size-ratios between neurocranium and splanchnocranium
 remains clear, as does the directional extension along the palate and
 the relative rotations from anterior to posterior and from caudal to cranial.

The situation is quite different for the comparison of the {\sl H. sapiens}
 average to our more distant relative, the female chimpanzee.  The 
 quadratic analysis analogous to Figure 16 can be found in Figure 18,
 but it no longer appears to look entirely like the bilinear map of Figure 14.
 Instead we encounter a strong {\sl local} feature of the transformation, the
 apparent flattening of the parietal region, that is seen in both
 of the thin-plate spline renderings of the top row (at left, for the
 actual shape coordinates; at right, for the quadratic fit) and likewise
 in the gridded representation of that quadratic fit at lower left.
 Strikingly, the residuals of this analysis seem no greater than those
 of the comparison of the {\sl sapiens} sample with the neanderthals,
 Figure 16, yet the flattening of the splines is clearly detected by this
 quadratic fit as well, which has so many fewer coefficients (and also a
 matrix inversion step of much lower rank, $5\times 5$ instead of
 $23\times 23$).  The bidirectional
 linearity of the lower left panel in Figure 16 has certainly 
 ceased to apply globally,
 while the hafting zone here seems still to be no sort of natural
 boundary between multiple modules.  The deformation remains          
 smoothly graded  except locally, in the parietal region.

 Yet when we switch the algorithm
 from the quadratic (five-term) fit to the cubic (nine-term) fit,
 Figure 19, nothing essential changes in the analysis as a result of
 these additional four degrees of freedom per coordinate.
 The thin-plate spline of the fitted points (upper right panel) is not
 much altered from that in the previous figure except in  that
 same nonconforming parietal region, and while the cubic fit here
 leads to pathologies of the extrapolated grid at every corner of the
 original scheme (lower left panel), its restriction to the interior
 of the actual anatomy, lower right in the figure, shows grid lines that,
 ignoring their curvature, are actually well-aligned with those of
 the lower left panel in Figure 16, the comparison
 from {\sl sapiens} to {\sl neanderthalensis.}
 We have thereby confirmed graphically that the shape difference in the
 parietal region is indeed {\sl local}.  Put this another way: the
 quadratic fit (Figure 18) and the cubic fit (Figure 19) convey the
 same message, a relatively continuous gradient of deformation right
 across the hafting zone.  And they agree, too, that the situation at
 the parietal (landmarks Opi through LaE)
 is {\sl not} coherent with this large-scale
 gradient. From Bregma forward, the lower right panels in Figures 17 and 19
 differ mainly in the intensity of rotation of these gridline
 segments; but posterior to that arbitrary boundary the parietal landmarks
 participate in a reorganization that is incommensurate between the two
 comparisons.

 Thus we see again that, just as in the Vilmann growth
 example, an approach that eschews all of the standard Procrustes steps 
 and also the usual thin-plate spline is capable of generating the
 same understanding of a morphological phenomenon, in this case a
 somewhat more complicated one. 

\vskip .2in

\centerline{\bf IV. Discussion}
\vskip .2in
        {\sl A. The main concern of GMM ought to be the transformation grid
 per se.} This was already clear from the earliest formal appearance
 of the concept in D'Arcy Thompson's {\sl On Growth and Form}
 (Thompson, 1917),
 where the review literature usually begins (even though
 portrait artists like Albrecht D\"urer had thought about this much
 earlier). The endpoint of the method ought to be not
 statistical but graphical, and the derived report should
 be geometrical, not statistical, en route to an ultimately
 biophysical or otherwise morphogenesis-informed endpoint. The main dilemmas
 in this tradition were
 already well-critiqued over the first six decades of its
 development as I reviewed them in Chapter 5 of Bookstein,
 1978. No matter how clearly defined the positions of individual
 landmark points might be, there was no complementary rhetoric for
 reporting meaningful features of the transformation grid that expressed
 comparisons of their configurations over meaningful biological contrasts.
 The best exposition of {\sl this} problem remains Sneath, 1967, a paper
 that struggled, ultimately unsuccessfully, to bring the algebra of
 landmark analysis (in that pre-spline era) into alignment with the
 reasoning of numerical taxonomy.
 Yet D'Arcy Thompson would have been delighted with the grid in
 Figure 13, while presentations of the same information in
 Procrustes style, Figure 1a, or spline-style, panels 4(a) through 11(a),
 would have been of no use to him at all.
 A more contemporary and quite distinct tradition of
 transformation studies approaches the problem via
 a calculus of {\sl diffeomorphisms}
 (see, for example, Grenander and Miller, 2007), which makes no essential
 reference to landmarks at all, instead basing its computations on the
 full field of image contents, gray-scale or even colored, spanning
 the organ(s) of interest.  The approach seems particularly helpful in
 neurological applications to imagery of the human brain.  This
 contrasting method, however, is beyond the scope of
 my Procrustes critique here.

        The analysis in Figure 7 suggests renewing Thompson's
 original concern in this domain, the interpretation of grids per se,
 via injecting a new theme into the discussion, an anatomical basis 
 for orienting the starting grid on the template,
 that more intensively exploits the interaction between deformation graphics and
 the investigator's prior awareness of how coordinate systems themselves
 can vary in their visually dominant features.
 The biomathematics ought to begin, then, with a confluence of
 two insights: one, that {\sl some}
 morphological domains might be amenable to {\sl some} kind of
 functionally interpretable large-scale pattern analysis, and the other,
 an intuition about the geometrical language by which the pattern
 of interest might be quantified.
 For Henning Vilmann, this translation began with the
 knowledge that growth of rodent neurocrania is a plausible domain for
 morphometric exploration and that its midsagittal aspect bears
 enough information about growth and function to be worthy of
 geometrization not only in his own measurements of extent, nor the numerous
 intermediate multivariate investigations
 of this same data set (including several of my
 own), but also in the novelties of Section II. But given these two axioms,
 an applied study would culminate in an exploration not of
 alternative {\sl statistics} but of alternative {\sl graphics}: a survey not of
 diverse linear combinations but of diverse grid renderings. 
 Information about absolute scale change, where relevant (as in
 biomechanical aspects of interpretation), can be
 embedded in any of these grid figures by a simple magnification over
 the course of printing, or can be inscribed on interlandmark segments
 or the line-elements of a transformation grid by overprinting.
 In this context of large-scale comparison, rotation is a tool of
 rendering clarification, not a nuisance variable of digitizing.

         The quadratic regressions in Figures 4 through 11
 all used the same list of five predictors
 $x, y, x^2, y^2, xy$.  This consistency lets
 the renderings here, unlike the approach in the lower row
 of Figure 1, preserve the uniform component of the transformation grid,
 where we can see how it interacts with these gradients of large but
 finite scale.  But the directions corresponding to those
 two axes $x$ and $y$ vary from baseline to baseline, and 
 the baseline points are not privileged by
 the regressions.  Consequently the coordinates pinned
 by the two-point registration are not quite pinned by the
 regression --- they are permitted to shift to
 some extent from solid to fitted circles
 in the grid figures here. 
 
        The resulting dataflow sheds new light on
 what we mean by ``{\sl the} best rotation'' when,
 as in both of this paper's examples,
 different parts of an organ appear to rotate
 relative to one another over a comparison of interest. 
 The role of the multiple two-point registrations that this paper recommends
 as a substitute for the Procrustes algorithm is 
 not itself a ``finding'' of any sort but merely a convenience,
 a simple way of
 regularizing the landmarks'
 Cartesian coordinates in order that a selection  of reasonable
 polynomial trends can be fitted, each in a reasonably equably weighted way.
 Its advantage is that unlike the case for
 the Procrustes method, there is more than one of them.
 The Procrustes approach optimizes a quantity 
 (sums of squares of landmark shifts)
 that is irrelevant to the ultimate purpose of an evolutionary
 or developmental GMM analysis, which is not a minimized sum of squares
 or a singular-value decomposition or a classification but rather
 a plausible biological hypothesis for
 the observed form-differences, their causes, or
 their consequences for the organism.
        
Then the logic of the inference engine
 we need is not the operationalized Procrustes arithmetic
 itself, the least-squares fit to what is almost always
 a completely wrong model (the null model,
 a pure similarity transformation).  Instead we need
 the logic of E. T. Jaynes's approach to numerical inference
 (e.g., Jaynes 2003): the explicit acknowledgement of what we do {\sl not}
 know --- what is missing from the list of
 data-driven constraints on some quantitative empirical inference.
 (I have recently reviewed this logic in the 
 rather different context of paleoseismology, which is the history of 
 great earthquakes --- see Bookstein 2021b.)
 What is missing from a Procrustes analysis is, among
 other things, the acknowledgement that choice of an orientation
 constraint affects the resulting report: what we seek is the orientation
 that will best clarify the final published diagram.  Furthermore, regardless
 of this issue of orientation, in every GMM 
 context we already know there is no ``correct'' registration, because
 there is no ``correct'' list of landmarks --- in the presence of
 any regional rotation
 or rescaling, different lists of landmarks or semilandmarks
 lead to different Procrustes registrations, and the empirical report of a
 shape comparison must accommodate that specific form of ignorance.
 That is the whole purpose of the grids --- to free our attention
 from  the landmark data per se to the space in-between where 
 biological processes actually take place.
   
       The particular protocol dictating the selection of
 orientations to be considered may be irrelevant to the 
 quantitative morphological inference under study.
 (Recall that in this paper
 the two points fixed in the baseline registration are
 not fixed by the fitted trend --- the registration is not
 an inferential component of the grid report at all.)
 Orientation may be specified as
 any interlandmark segment from  the available pairings,
 or any homologous boundary alignment, or even a specific force
 vector such as a muscle load or gravitational vertical ---
 or possibly all of these.
 Whatever the choices of orientation, the investigator of
 a global deformation is led
 to the approach here, which is the selection of at least one satisfactory
 such orientation as judged by the ultimate diagram at the end of
 the workflow.  In 3D, one could proceed via an assortment of
 large landmark triangles passing near the centroid,
 similarly searching for clarity and redundancy.
 But in other contexts that issue of orientation may be
 quite relevant to the interpretation. The examples here have all
 dealt with global trends, but Figures 18 and 19 hinted at
 a need for a deformation tool suitable for local features as well.
 Such a tool would likewise entail a rotation of the Cartesian coordinate
 system prior to grid computation, but in general a different
 one --- see, for example, the model of the
 {\sl crease} in Bookstein 2000 or Bookstein 2014, Figure 7.19.

\vskip .2in
         {\sl B.  We need to broaden the range of ideas we borrow
 from geometry.}
 A combination of two branches of geometry led us to the bilinear
 interpretation in Figure 14 of
 the grid in Figure 13, but this other toolkit is not among
 those currently
 being taught to biomathematicians. The kernel $r^2~\log~r$
 of the thin-plate spline doesn't much resemble the biological 
 processes we are trying to understand, but the algebra of 
 polynomial fits (here, mainly the specific appearance of
 bilinear maps leaving both pencils of coordinate lines almost
 straight and almost evenly spaced after deformation) does pick up
 much of the classic appearance of growth-gradients as laid
 out for analysis from Thompson on. More important than the
 extension of the idea of a coordinate system, though, is an
 extension of the domain of morphometric data to include empirical
 entities other than landmark points.
 The description of
 the grid in Figure 13 makes no essential mention
 of {\sl any} of the landmarks ---
 the simple exegesis here (bilinear reorganization of that
 particular family of grid lines while
 remaining lines) pertains much more to the interior of this octagon 
 (the directions of those transects across it, or, if you will, the pairing of
 points across the left and right sides of the outline in this orientation) than
 to any of its boundary delineation detail,
 even though that boundary is the sole data source
 for the example. Thus at root the finding exemplifies a language of
 intraorganismal matching, the pairing of points along a shared curve
 bounding some anatomical entity in section.
 Pairings like these are not like landmarks in any formal aspect.
 
So even though this paper's first example argument began
 from a playful GMM-derived diagram, Figure 1d, it ends up formalized in the
 rhetoric of a spatial extension (Figure 13) unknown to GMM but comprehensible
 by every reader of Thompson's chapter, as
 interpreted in Figure 14 via a similar-looking figure from a subchapter of
 college geometry.  This logical sequence can be reversed: beginning
 from those same textbooks, to try finding biological examples that illustrate
 them. We are used to polar coordinates, for example (most recently in
 the study of centric allometry, Bookstein 2021a), but what about 
 bipolar coordinates or confocal coordinates
 (Bookstein 1981, 1985) and other schemes
 that (literally) co-ordinate position with respect to two origins or two
 axial systems at the same time?  The range of coordinate systems is vastly
 broader than the Cartesian on which today's GMM automatically relies.
 My biorthogonal grids (Bookstein 1978) already went beyond this
 possibility, though not in a statistically feasible way,
 via their formalism of one-axis and three-axis singularities corresponding
 to the ``lemon'' and ``star'' umbilics that are the topic of
 advanced treatises such as Koenderink (1990).
 From the earliest years of the twentieth century the
 mathematics of geometry has permitted us to talk
 about coordinates of many different extended structures:
 not just points, but lines, planes, circles,
 and many other formalisms.  See, at first,
 Hilbert and Cohn-Vossen, 1931/1952, and then, among the more
 contemporary surveys, Porteous 2001 or Glaeser 2012.

       Thus the word ``geometric'' in the phrase ``geometric morphometrics'' needs
 to have its meaning broadened beyond the current focus on
 the Procrustes component of GMM or indeed any version based on
 analysis of landmark points as logically separate data elements. 
 ``Procrustes distance'' between
 specimens, when computed as a minimizing sum of squared Cartesian
 coordinate differences, is just a theory-free
 proxy for the far more subtle and multifarious concept the biologist knows as
 the opposite of ``similarity,'' and today's
 GMM treats Procrustes shape coordinates 
 as just a list of Cartesian pairs (or triples) in their own coordinate space of
 position, without reference to any explicit features for describing how their
 interrelationships (e.g. the interlandmark segments of Figure 1) actually
 change across a comparison of configurations.
  D'Arcy Thompson got this correct back in 1917:
 ``The deformation of a complicated figure,'' he wrote (Thompson 1961:271),
 ``may be a phenomenon
 easy of comprehension, though the figure itself have to be left unanalyzed and
 undefined.  This process of comparison, recognizing in one form a definite
 permutation or deformation  of another, apart altogether from a precise
 and adequate understanding of the original `type' or standard of comparison,
 lies within the immediate province of mathematics.''

 That geometry of
 ``recognizing deformation'' is not limited to the geometry of points referred
 individually to Cartesian axes. Thompson himself referred explicitly
 to the appearance of the deformed grid lines in his drawings.
 For the comparison to {\sl Mola,} for instance, he wrote,
 ``I have deformed [{\sl Diodon's}] vertical coordinates into a system
 of concentric circles, and its horizontal coordinates into 
 a system of curves which, approximately and provisionally, are
 made to resemble a system of hyperbolas'' (Thompson 1961:300). It
 is the configuration of these curves, not the landmarks on them, that
 is the bridge from arithmetic to understanding.
 In other words, the elementary language of 
 deformation, the language by which we report morphological
 comparisons {\sl as} deformations,
 must be based in a glossary of multiple elementary types of deformable 
 image components, not disarticulated
 landmarks.  The roster of these is broad indeed, including, among
 other options,
 the changes of point-pairs to other point-pairs at a different distance
 or direction that we already saw in Figure 1, but also changes of
 triangles to other triangles, squares to any 
 quadrilateral whether rectangle, parallelogram,
 trapezoid, or some other form,
 displacement of interior points with respect to an unchanging
 boundary, circles to ellipses, ellipses to
 any other simple closed curve, straight lines to other straight
 lines, lines to any other
 open curve, line-elements having an orientation in the small
 as well as a location (for a spline cognizant of this
 structure, see Bookstein and Green, 1993),
 or nearby pairs of parallel lines to any bent ribbon tracing the sequence of
 changes all along their shared length.  
 All of these have appeared in biometric examples; each requires
 a different geometric grammar for its reporting.  For instance
 (in another acknowledgement of our sister discipline
 of neuromorphometrics), line elements per se summarize image data for
 the method known as diffusion tensor analysis that traces and
 summarizes patterns of wiring in the human brain.
 
As I hope you have already come to suspect from the figures in this paper,
 the thin-plate spline is not designed to be
 of any particular help in this matter.  Its functional form is mainly a sum of terms
 $r^2 \log r,$ where $r$ is the distance from each grid point to each
 landmark of the template in turn, and so it has no machinery for collecting references
 to two or more landmarks at the same time, but must revert to the nonbiological
 symmetries of linear multivariate statistics for this purpose (so that
 the partial warps, for instance, are just a $(2k-4)-$dimensional
 rotation of its Cartesian coordinates however they were arrived
 at to that point, while the relative warps are just a different
 $(2k-4)-$dimensional rotation of the same coordinates).
 No, the elements of a quantitative morphometric comparison
 in terms of deformation must be the
 whole coordinate systems of our deformation diagrams, and the features we
 extract must be features that refer to
 those deformed lines and areas, whether end to end
 or truncated to the vicinity of specific landmark subsets.
 Any geometric report qualified to drive a programme 
 like Thompson's aimed at simple descriptions of relationships among
 individually complicated specimens must
 begin with more complicated elementary entities than positions of discrete landmark
 points. A search for such explananda, beginning from the paired interlandmark segments in
 Figure 1, leads immediately to the elementary aspects of this paper's two examples,
 which make no reference to the formula $r^2 \log r$ nor indeed any quantification
 beyond the squaring or cubing of coordinates and products of those powers that
 allows us to parameterize families of nearly parallel curves
 that began as parallel lines.
\vskip .2in

         {\sl C.  The exterior of an organ or an organism is a useful domain
 for communication of findings even in the absence of tissue.}
 This comment has real bite for a GMM that depends on the conventional
 thin-plate spline, which does not understand exteriors at all.
 So the usual interpolating spline is precisely
 the wrong tool for detecting large-scale
 gradients that, like the one summarizing
 the Vilmann comparison, are not affine ---
 are not conducive to descriptions emphasizing some pair of directions
 at $90^\circ$ bearing the maximum
 ratio of rates of change.   Because the conventional thin-plate spline
 relaxes to uniform at great distances, it is not a helpful
 component of answers to any question about large-scale organization of
 a form-comparison, the question asked
 by most morphologists (and dysmorphologists,
 and paleontologists) ever since Thompson's time.
 To quantify the cunning hint from Figure 1d, I needed the
 tool of a quadratic trend surface (i.e., a fit, not an interpolation), 
 and when the graphic of that fit proved intriguing, a suitable
 summary arose only when the rendering was extended (Figure 13) far enough
 beyond the actual convex hull of the landmarks that Figure 14 could show us how
 to report its structure. However vague the language might be for a discussion
 of Figure 7 by itself, the reworking that is
 Figure 13 makes the implicit explicit ---
 the extended grid now is exactly the report we seek,
 no actual words required except the legend explaining how the graphic
 was produced. But such a graphic no longer resembles any sort
 of conventional GMM output.

        Because the interior of any non-nested module is at the same time a
 part of the exterior of every other module, one sees from the hominization
 example that the morphometric aspect of
 ``modularity,'' whatever its exact morphogenetic definition,
 is a matter not of landmark coordinates but of what
 happens to coordinate grid lines.   
 Figures 17 and 19 confirm that, within the limits of these data
 resources (adult forms only, no growth series, a mere 20 landmarks),
 there is no graphical
 evidence for the cranial base as a separatrix  between braincase and
 face, in spite of their obvious differences in function, but strong
 evidence for a separation of the whole anterior two-thirds of this
 landmark scheme from the five parietal landmarks, Opi through LaI and LaE,
 that so clearly seize control of the lower-right corner
 of the grids for either the quadratic fit (Figure 18 lower left) or
 the cubic fit (Figure 19 lower right) to the comparison
 across genera. While the empirical import of
 this second data example is obsolete, owing to advances in the accrual of
 samples of all these species, the practice whereby consideration of the
 transformation grids per se might shape inferences from  landmark data
 about morphogenetic control processes ought to be transferred from
 the current GMM toolkit to these more integrated investigative tools
 along the lines of the examples here.

\vskip .2in
       {\sl D.  The implications of a diminished role for
 the existing core of geometric morphometrics
 in quantitative morphology are liberating.}
 Via a new toolbox that intentionally discards Procrustes
 centering, Procrustes scaling, and Procrustes orientation,
 and that downplays the role of
 thin-plate splines --- the whole core of today's GMM ---
 we may be able to better achieve GMM's principal declared purpose,
 the quantitative understanding of morphological variation and its causes
 or effects, by recourse to more diverse geometrical formalisms, some ancient
 and some relatively novel.  
 This methodological possibility has several
 implications, some for actual analysis of morphologies and others for
 the methodological component of graduate
 curricula in the evo-devo sciences.  The aspects of
 geometry that GMM is accustomed to borrowing for its tools concentrate much too
 heavily on matrix algebra and linear multivariate analysis. As Peter
 Sneath suspected so long ago in his paper on trend-surface analysis, there are
 other geometric entities, such as those here dealing with quadratic 
 bivariate polynomials, that speak more
 clearly to the investigator's visual instincts,
 especially as regards phenomena of orientation. (Examine, for instance,
 panel 1d of Sneath 1967,\footnote{$^5$}{According to Biegert 1957,
 the orientation is along the central plane of the sphenoid
 (in Latinate German, ``Planum-sphenoideum-Ebene'') to suit the needs of a much
 broader study of the midsagittal skull across the order {\sl Primates}.}
 which shows a relative rotation between face
 and braincase in the comparison of {\sl Homo} to {\sl Pan}
 similar to the
 one in Figure 18 here, without, however, the optimization  of coordinates
 that Section III exploited.) And far more objects
 can be assigned coordinates than discrete points (or semilandmarks) alone:
 grid lines, for instance, deserve coordinates
 of their own (Figures 4 through 11)
 and also interlandmark segments (Figure 1).  

       Similarly, the way GMM relies on thin-plate splines for its
 published renderings  exaggerates their importance
 for organismal biology. The spline is an interpolating map, whereas,
 in view of how arbitrary our landmark lists actually are,
 biological interpretation often goes deeper and better via
 approximating maps instead.  The actual role of interpolating splines in
 the research cycle, then, might be shifted well earlier,
 all the way back to before the final rendering style is chosen,
 in order to supply guidance 
 about which geometrical languages should be exploited for the most effective
 dissemination.  At that early stage,
 interpolating splines are good aids to the 
 search for component processes that are primarily local,
 but are poor at the analogous global reports, which, as Sneath
 already knew in 1967, do better with polynomial analyses.  Both
 possibilities should be checked, and perhaps both preserved in
 the final analysis, the way Figures 4 ff. show both the thin-plate
 spline, which reveals the local change at IPS, and the quadratic
 grid, which summarizes the overall change of form so much better
 (in both contexts ignoring the Procrustes side of GMM in favor
 of the different optimization of orientation recommended here).

 The finding in Figure 1d
 should not have been new to this paper. In the many previous GMM
 investigations of the Vilmann data there should long since have been mention
 of rotations of subanatomies, a rhetoric that has been suppressed,
 perhaps unintentionally, by virtue of 
 our current traditions of overly symmetric data summaries like 
 Procrustes distance, principal
 component analysis and interpolating splines.\footnote{$^6$}{In an
 ironic exception, a non-Procrustes analysis in my
 1991 textbook refers to this rotation as an epiphenomenon (a side-effect) of
 {\sl orthocephalization}, the usual name for the process by which
 the anterior cranial base thrusts under the facial skeleton
 --- but the verb ``rotate'' itself is in
 scare quotes!  See Bookstein (1991:312).} It is time for
 the morphological side of biomathematics to return to its roots in biological
 geometry {\sl sensu lato} --- what might the organism's function space
 ``know'' about its own form? --- in order to rebuild the interplay between data
 and explanation using a much broader range of geometric formalisms than
 just ``points'' (or their ``modules'')
  and ``deformations.'' The method of cubic regression, Figures 17 and
 19, is likewise not new; I copied it straight from Sneath (1967).
 The particularly careless way the Procrustes method
 dismisses orientation as just a nuisance variable has blinded our field to
 the possibility that relative intraspecimen orientations
 can be just as informative a channel of insight
 and explanation as relative extents (proportions). 
 To restore and then extend this symmetry we need to abandon the
 standard Procrustes tool in favor of explorations that explicitly consider
 multiple orientations at the same time, just as studies of allometry have
 been considering multiple size measures
 since at least Blackith and Reyment (1971).
 More generally, to understand transformation grids we must
 extend our understanding of the sort of entities that can {\sl have}
 coordinates from points to more extended structures. Only then
 can we trust our diagrams to provide straightforward
 practical summaries of the ``blooming, buzzing confusion'' (W. James) that
 is the spectrum of Darwinian phenomena we call evo-devo.

\vskip .3in
{\bf Acknowledgements.} I thank Jim Rohlf, Stony Brook University, for
 thoughtful commentary on the basic thrust of this manuscript at several
 earlier stages. It was Joe Felsenstein, University of Washington,
 who first alerted me to foundational problems in the way
 GMM handles the concept of ``rotation.'' 
\vskip .1in
{\bf Competing interests and funding.}
 There has been no support from any external funding
 source, and no conflicts of interest thereby. 
\vskip .1in

\vfill\eject
 
\centerline{\bf Literature Cited}
\vskip .2in

Biegert, J.  Der Formwandel des Primatensch\"adels.
 {\sl Gegenbaurs morphologisches \hfill\break Jahrbuch} 98:77--199, 1957.

Blackith, R. E., and R. A. Reyment
 {\sl Multivariate Morphometrics.}
 Academic Press, 1971.

Boas, F.  The horizontal plane of the skull
 and the general problem of the comparison
 of variable forms.  {\sl Science} 21:862--863, 1905.

Bookstein, F. L.  {\sl The Measurement of Biological Shape and Shape
      Change}.  Lecture Notes in Biomathematics, vol. 24.  Springer-Verlag,
      1978. 

Bookstein, F. L.  Coordinate systems and morphogenesis.  In
      {\sl Morphogenesis and Pattern Formation}, ed. T. G. Connelly,
      L. Brinkley, and B. Carlson. Raven Press, 1981, pp. 265--282.

Bookstein, F. L.  Tensor biometrics for changes in cranial
      shape.  {\sl Annals of Human Biology} 11:413--437, 1984.

Bookstein, F. L.  Transformations of quadrilaterals, tensor
      fields, and morphogenesis.
      In {\sl Mathematical Essays on Growth and the
      Emergence of Form}, ed. P. L. Antonelli.  University of
      Alberta Press, 1985, pp. 221--265.

Bookstein, F. L.  Size and shape spaces for landmark data in
      two dimensions.  (With Discussion and Rejoinder.)
      {\sl Statistical Science} 1:181--242, 1986.

Bookstein, F. L.
  {\sl Morphometric Tools for Landmark Data: Geometry and Biology}.
      Cambridge University Press, 1991. 

Bookstein, F. L., and W. D. K. Green.  A feature space
 for edgels in images with landmarks.  {\sl Journal of Mathematical
 Imaging and Vision} 3:231--261, 1993.

Bookstein, F. L. Creases as local features of
 deformation grids. {\sl Medical Image Analysis} 4:93--110, 2000.

Bookstein, F. L., P. Gunz, P. Mitter\"ocker, H. Prossinger,
 K. Sch\"afer, and H. Seidler.  Cranial integration in
 {\sl Homo}: Singular warps analysis of the midsagittal
 plane in ontogeny and evolution.  {\sl Journal of
 Human Evolution} 44:167--187, 2003.

Bookstein, F. L., and P. M. Mitteroecker.
 Comparing covariance matrices by relative eigenanalysis, with applications
 to organismal biology.  {\sl Evolutionary Biology} 41:336--350, 2014.

Bookstein, F. L. {\sl Measuring and Reasoning: Numerical Inference
 in the Sciences.}  Cambridge University Press, 2014.

Bookstein, F. L. Reconsidering ``The inappropriateness of
 conventional cephalometrics.''
 {\sl The American Journal of Orthodontics} 149:784--797, 2016.

Bookstein, F. L.  A method for factor analysis of shape
 coordinates. {\sl American Journal of Physical Anthropology}
 64:221--245, 2017.

Bookstein, F. L.  {\sl A Course of Morphometrics
 for Biologists.}
 Cambridge University Press, 2018.

Bookstein, F. L. Centric allometry:
 Studying growth using landmark data.
 {\sl Evolutionary Biology} {\tt doi:/10.1007/s11692-020-09530-w,}
 48:129--159, 2021a.

Bookstein, F. L. Estimating earthquake probabilities by
 Jaynes's method of maximum entropy. {\sl Bulletin of
 the Seismological Society of America} 111:2846--2861, 2021b.

Claude, J.  {\sl Morphometrics with R.}  Springer, 2008.

Galton, F.  Classification  of portraits.  {\sl Nature} 76:617--618, 1907.

Garson,  J. G.  The Frankfort craniometric agreement, with
 critical remarks thereon.  {\sl Journal of the the Anthropological
 Institute of Great Britain and Ireland}, 14:64--83, 1885.

Glaeser, G.  {\sl Geometry and its Applications
 in Arts, Nature, and Technology.} (English edition, modified
 from the German original.) Springer, 2012.

Grenander, U., and M. Miller.
 {\sl Pattern Theory: from Representation to Inference.}
 Oxford University Press, 2007.

Hilbert, D., and D. Cohn-Vossen.  {\sl Anschauliche
 Geometry.} 1931.  Tr. by P. Nemenyi as {\sl Geometry and
 the Imagination,} Chelsea Pub. Co., 1952.

Jaynes, E. T.   {\sl Probability Theory: the Logic of Science.}
 (Ed. G. L. Bretthorst.)  Cambridge University Press, 2003.

Koenderink, J.  {\sl Solid Shape.}  MIT Press, 1990.

Martin R.   {\sl Lehrbuch der Anthropologie
 in systematischer Darstellung.} Jena: Gustav Fischer, 1914.

Porteous, I. R.  {\sl Geometric Differentiation for the
 intelligence of Curves and Surfaces,} 2nd ed.
 Cambridge, 2001.  

Przibram, H. {\sl Form und Formel im Tierreiche. Beitr\"age zu einer
 quantitativen Biologie I--XX.} Franz Deuticke, 1922.

Sneath, P. H. A.  Trend-surface analysis of transformation grids.
 {\sl Journal of Zoology, London} 151:65--122, 1967.

Stuart, A., and K. Ord.
 {\sl Kendall's Advanced Theory of Statistics.}
  Volume 1, {\sl Distribution Theory.}
 Wiley, 1994.

Thompson, D'A. W.  {\sl On Growth and Form.} Macmillan, 1917.  Abridged
 edition, ed. J. T. Bonner, Cambridge University Press, 1961.

Weber, G. W., and F. L. Bookstein.  {\sl Virtual Anthropology:
 a Guide to a New Interdisciplinary
 Field.} Springer Verlag, 2011.

\vskip .2in
\centerline{\bf Captions for figures}
 
\vskip .1in
{\bf Figure 1.} Unexpected pattern in the much-analyzed
 Vilmann data set of neurocranial octagons for growing laboratory
 rats.  (upper left) Saturated network of interlandmark segments,
 Procrustes average shapes of the octagons at ages 7 days (light lines)
 and 150 days (heavy lines). Landmarks: Bas, Basion; Opi, Opisthion;
 IPS, Interparietal suture; Lam, Lambda; Brg, Bregma; 
 SES, Spheno\"ethmoid synchondrosis; ISS, Intersphenoidal synchondrosis;
 SOS, Spheno\"occipital synchondrosis.  (upper right) Subnetwork of segments
 rotating by at least 0.15 radians ($8.6^\circ$) over this age comparison.
 (lower left) The same saturated network for the nonaffine component
 only of the same Procrustes shape coordinates,
 with landmark numbers.  (lower right)
 Now tht the uniform component of this shape coordinate
 space has been partialled out, there emerges a considerably
 simpler subnetwork, explicitly displaying the relative
 rotation of the anteriormost three landmarks with respect to the other five. 
\vskip .1in
{\bf Figure 2.} Two-point superpositions (Bookstein coordinates)
 of the Vilmann age-7 and age-150 average octagons for every possible
 baseline. Landmarks are numbered as in Figure 1.  Circled landmarks:
 ends of the baseline as registered to $(0,0)$ and $(1,0)$.  Light lines,
 age-7 average; heavy lines, age-150 average.  
\vskip .1in
{\bf Figure 3.} Contrasting morphometric renderings for
 diverse transformations by thin-plate spline (lower row) of a variously
 oriented square (upper row).  (a) Square to parallelogram,
 grid aligned with the edges of the square.
 (b) The same, grid now aligned with the square's diagonals.
 (c) Square to trapezoid.  (d) Almost the same, grid rotated $45^\circ$:
 square to kite.  Adapted from Bookstein, 1991, Figure 7.3.6.
\vskip .1in
{\bf Figure 4.} This is the first of eight figures that
 all have the same four-panel format as applied to one of eight
 selected baselines from the array of 28 offered in Figure 2.
 Top-row panels, left to right: actual change of averaged
 Cartesian coordinates, with thin-plate spline oriented to selected
 baseline; ordinary thin-plate spline of the quadratic fit to the
 age-150 average as regressed on first and second powers
 of the $x-$ and $y-$ coordinates of the
 template and also their product $xy$.
 Bottom row, left, the quadratic fit (not a spline) as a
 grid of its own.  Solid circles, the
 observed data; open circles, predictions from this regression.
 Bottom right: restriction of the display list of
 grid vertices to the interior of the age-7 octagon as explained in the
 text.  The baseline of this figure runs from Basion to Opisthion (landmark
 1 to landmark 2).
\vskip .1in
 {\bf Figure 5.}  The same as Figure 4 for a baseline
 from Basion to Interparietal suture, landmark 1 to landmark 3.
\vskip .1in
 {\bf Figure 6.}  The same for a baseline
 from Basion to Lambda, landmark 1 to landmark 5.
\vskip .1in
 {\bf Figure 7.}  The same for a baseline
 from Interparietal suture to Spheno\"occipital synchondrosis, landmark 3 to landmark 8.
 Each panel is roughly a $90^\circ$ rotation of the corresponding panel in
 Figure 6, having a baseline at about $90^\circ$ to this one.
\vskip .1in
 {\bf Figure 8.}  The same for a baseline
 from Lambda to Spheno\"occipital synchondrosis, landmark 4 to landmark 8.
\vskip .1in
 {\bf Figure 9.}  The same for a baseline
 from Bregma to Spheno\"ethmoid synchondrosis, landmark 5 to landmark 6.
\vskip .1in
 {\bf Figure 10.}  The same for a baseline
 from Bregma to Intersphenoidal suture, landmark 5 to landmark 7.
\vskip .1in
 {\bf Figure 11.}  The same for a baseline
 from Spheno\"ethmoid synchondrosis to Basion, landmark 6 to landmark 1.
\vskip .1in
 {\bf Figure 12.}  Synthesis of upper left and lower right panels
 of Figures 4 through 11, analyses to eight of the 28 possible two-point
 baselines. Clearly some of these choices lead to simpler reports
 than others do. As the thin-plate spline is covariant with similarity
 transformations of its target, all the splines here (columns 1 and 3)
 are the same except for grid orientation and spacing.  But the regressions
 associated with columns 2 and 4 weight different landmarks
 differently (in particular, weighting the two ends of the baseline
 not at all), so these grids can vary in more aspects than the
 baseline orientation per se.
\vskip .1in
 {\bf Figure 13.} Graphical extension of the quadratic fit to the IPS-SOS baseline
 yields a striking reinterpretation of the phenomenon.  Left, grid extended to
 the left over the baseline-registered template; right, corresponding
 version of the fitted quadratic trend from Figure 7.
 Filled dots, observed average configurations after the two-point
 registration (left, age 7 days; right, age 150 days). Open dots, 
 fitted values of the quadratic regression as in the earlier figures.
\vskip .1in
{\bf Figure 14.} Two alternatives for column (d) of Figure 3.
 The map in Figure 13 more closely resembles the
 bilinear map (far right panel) than the projection map (central
 panel).  The projection map sends {\sl all} straight lines to other
 straight lines; the bilinear map, in general, only the lines that
 join matched proportional aliquots from opposite edges.
 In both deformations the dashed line delineates the effect of the map
 on the horizontal diameter of the starting diamond shape.  The projection
 takes this curve to a straight line, the bilinear map, to a parabola
 engendering a less extreme reduction of the 
 template cells' areas above this diameter.
 Owing to the shared symmetry axis of square and kite
 there is another set of straight lines within the grid
 in the rightmost panel --- the
 verticals --- but this third set is not present in the general
 case, hence the ``bi'' of ``bilinear,'' and so I have not
 drawn them here.
\vskip .1in
{\bf Figure 15.} Landmark configurations for the hominization example,
 Section III.  (left) Abbreviated names of the twenty landmarks printed
 at the raw digitized coordinate averages of the
 adult {\sl sapiens} subsample 
 of Bookstein et al. (2003). 
 Alv, alveolare, inferior tip of the bony septum between the two
maxillary central incisors;
 ANS, anterior nasal spine, top of the spina nasalis anterior;
 Bas, basion, midsagittal point on the anterior margin of the foramen magnum;
 BrE, BrI, external and internal bregma, outermost and innermost
 innermost intersections of sagittal and lambdoidal sutures;
 CaO, canalis opticus intersection, intersection point of a chord connecting
 the two canalis opticus landmarks with the midsagittal plane;
 CrG, crista galli, point at the posterior base of the crista galli;
 FCe, foramen caecum, anterior margin of foramen caecum in the
midsagittal plane;
 FoI, fossa incisiva, midsagittal point on the posterior margin of the fossa incisiva;
 Gla, glabella, most anterior point of the frontal in the midsagittal;
 InE, InI, external and internal inion, most prominent projections of the
 occipital bone in the midsagittal;
 LaE, LaI, external and internal lambda, outermost and
 innermost intersections of sagittal and lambdoidal sutures;
 Nas, nasion, highest point on the nasal bones in the midsagittal plane;
 Opi, opisthion, midsagittal point on the posterior margin of the foramen magnum;
 PNS, posterior nasal spine, most posterior point of the spina nasalis;
 Rhi, rhinion, lowest point of the internasal suture in
 the midsagittal plane;
 Sel, sella turcica, top of dorsum sellae;
 Vmr, vomer, sphenobasilar suture in the midsagittal plane.
 (right) Bookstein coordinates to an ANS-LaI baseline
 for the averaged adult {\sl H. sapiens} and
 {\sl H. neanderthalensis} samples and the single adult female chimpanzee. 

\vskip .1in
{\bf Figure 16.}
     Three grid diagrams for the comparison of the averaged {\sl H. sapiens}
 and {\sl H. neanderthalensis} twenty-landmark configurations, to an ANS-LaI
 baseline.  (upper left) Conventional thin-plate spline grid deforming the
 {\sl sapiens} average to the {\sl neanderthalensis.}  (upper right) Thin-plate
 spline rendering of the deformation from the same averaged {\sl sapiens} to
 the quadratic regression fits (regressions on first and second
 powers of the $x-$ and $y-$coordinates and also their product $xy$)
 of the {\sl neanderthalensis} configuration.
 (lower left) Explicit grid of that quadratic regression. Solid circles,
 observed averaged {\sl neanderthalensis} two-point coordinates; open
 circles, fitted locations.

\vskip .1in
{\bf Figure 17.} The same for a cubic regression of the {\sl neanderthalensis}
 coordinates, nine predictors instead of five. Upper left, upper right,  and
 lower left panels as in Figure 16.  At lower right, an enlarged version of
 the fitted grid (lower left) as trimmed to the interior of the actual
 {\sl neanderthalensis} average.

\vskip .1in
{\bf Figure 18.}  The same as Figure 16 for the comparison of the averaged
 {\sl H. sapiens} to the single female chimpanzee in the data base of
 Bookstein et al. 2003.  

\vskip .1in
{\bf Figure 19.}  The same as Figure 18 for the comparison of {\sl H. sapiens}
 to the female {\sl Pan} using these tools.  The grid at lower left, for
 the cubic fit, is correctly drawn even though it looks like a whale.

\vfill\eject

\centerline{\epsfxsize=6truein\epsffile{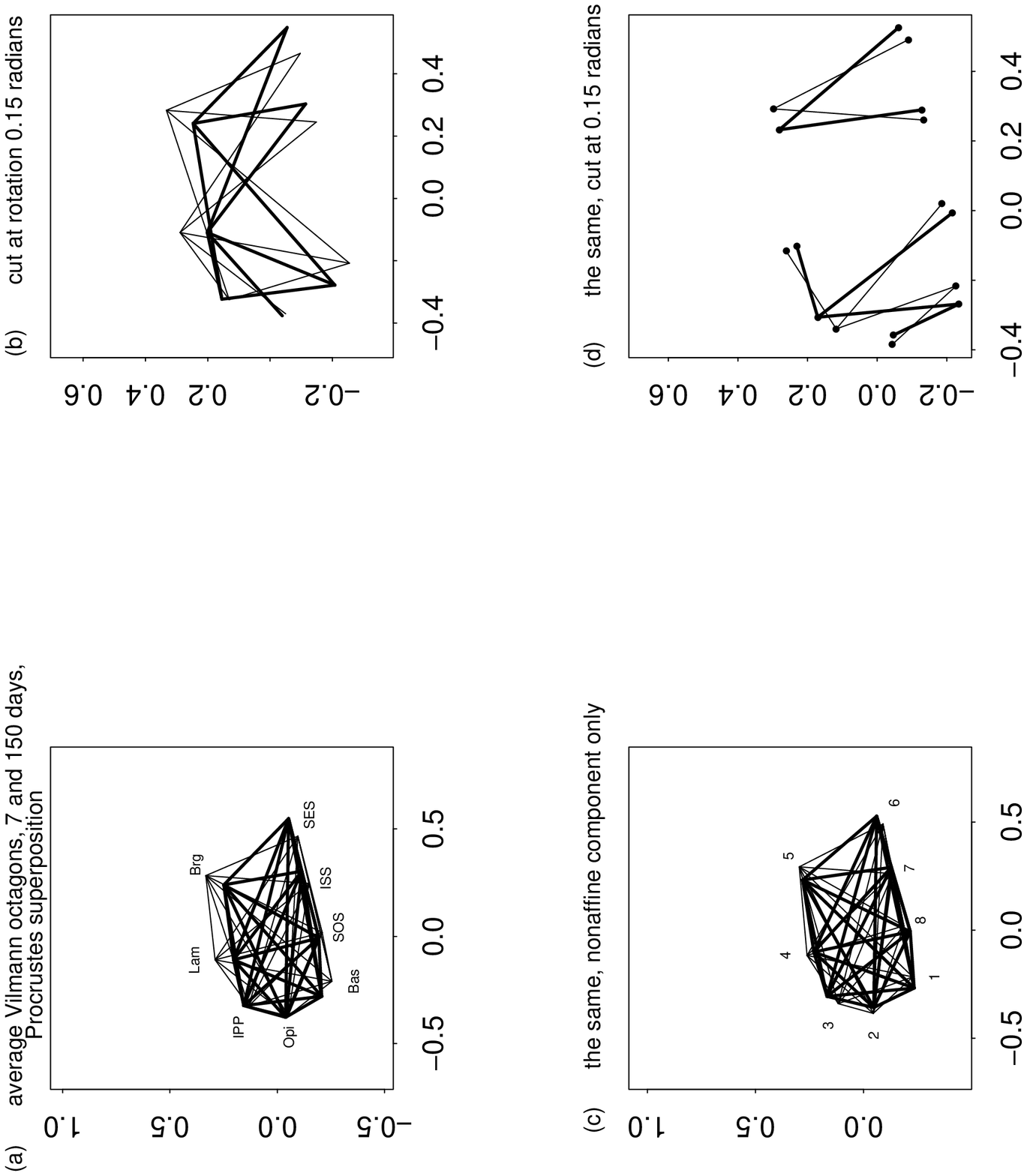}}
\vskip .2in
\noindent{\bf Figure 1.}
\vskip .2in
\centerline{\epsfxsize=6truein\epsffile{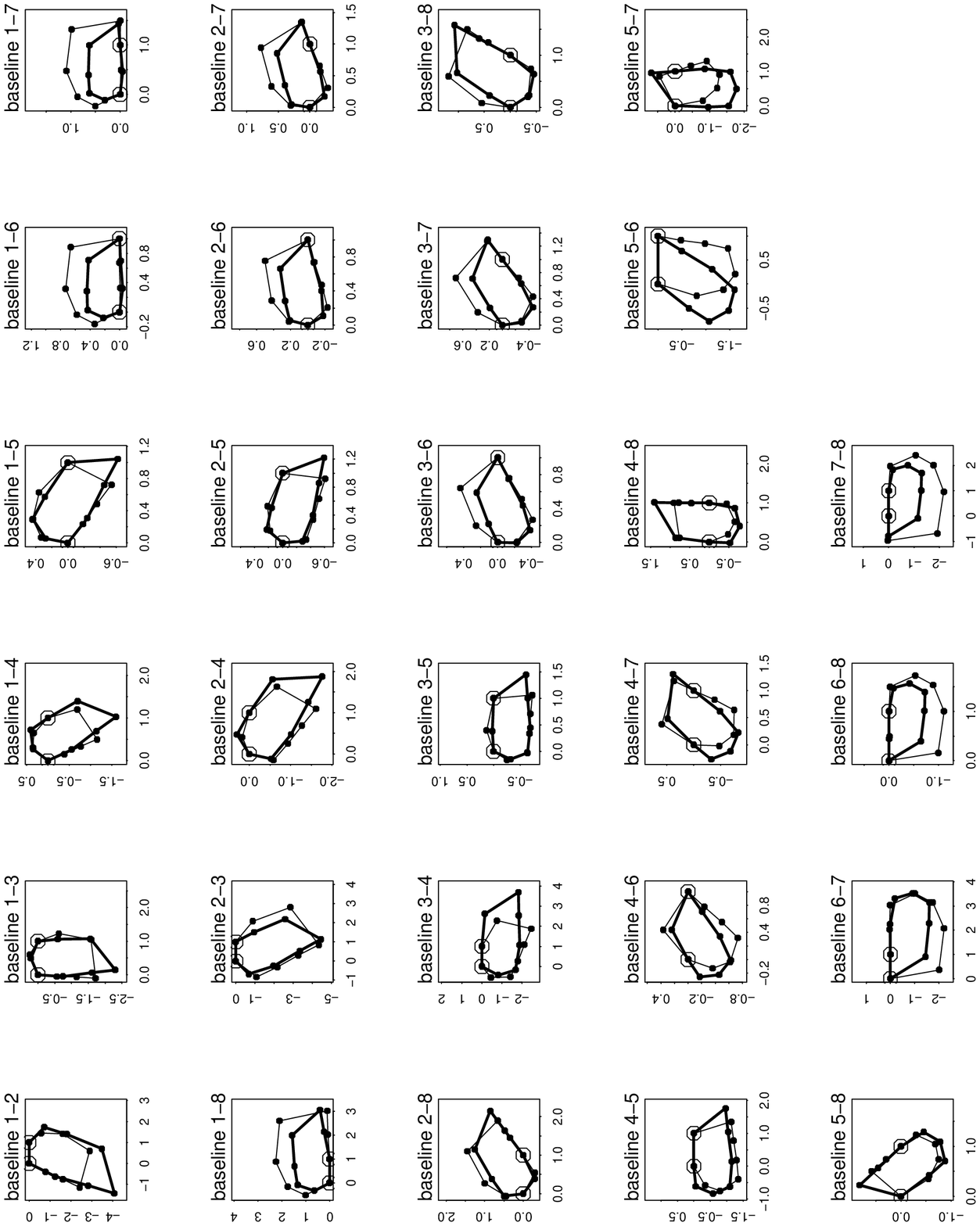}}
\vskip .2in
\noindent{\bf Figure 2.}
\vskip .2in
\centerline{\epsfxsize=6truein\epsffile{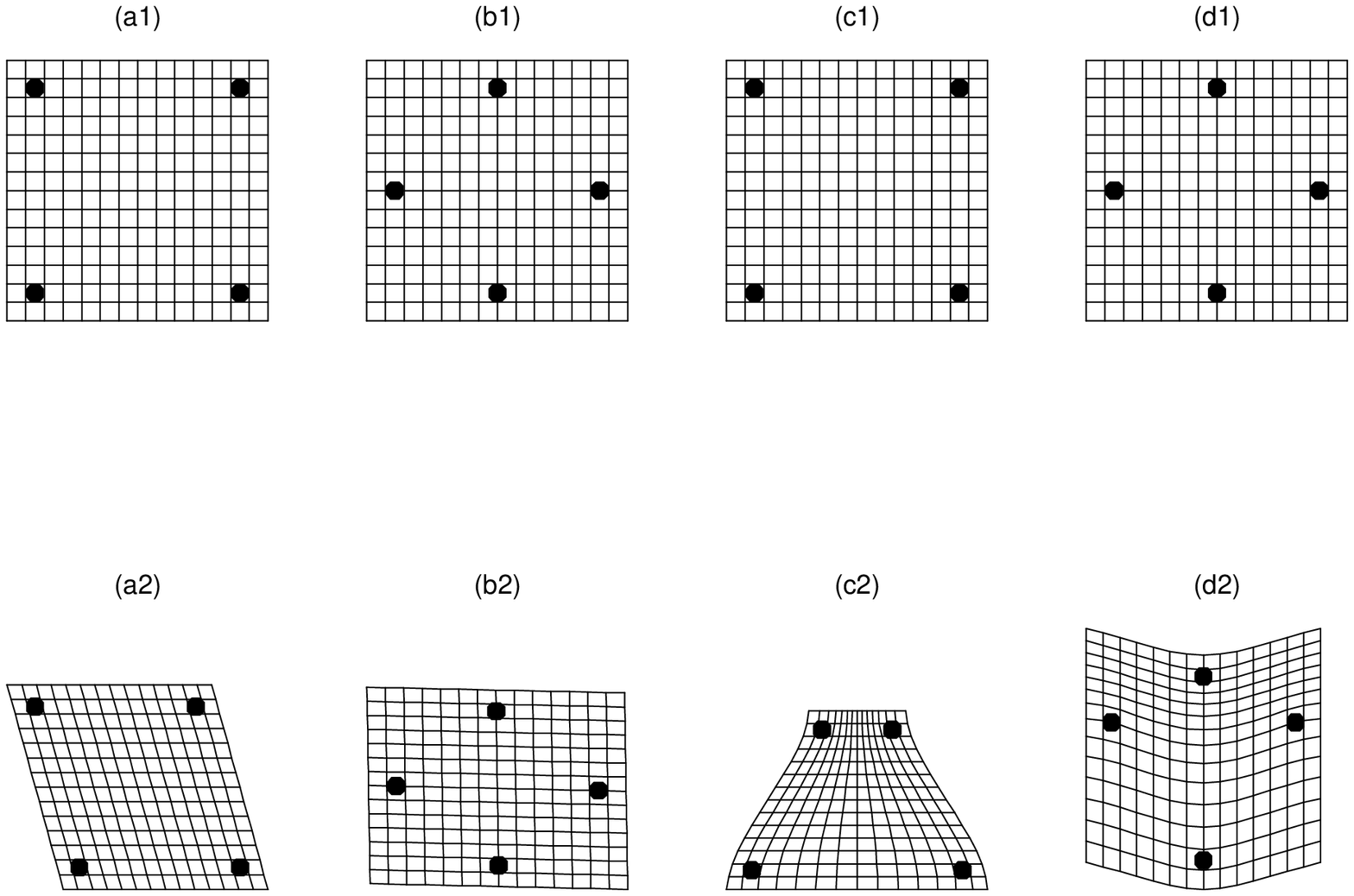}}
\vskip .2in
\noindent{\bf Figure 3.}
\vskip .2in
\centerline{\epsfxsize=6truein\epsffile{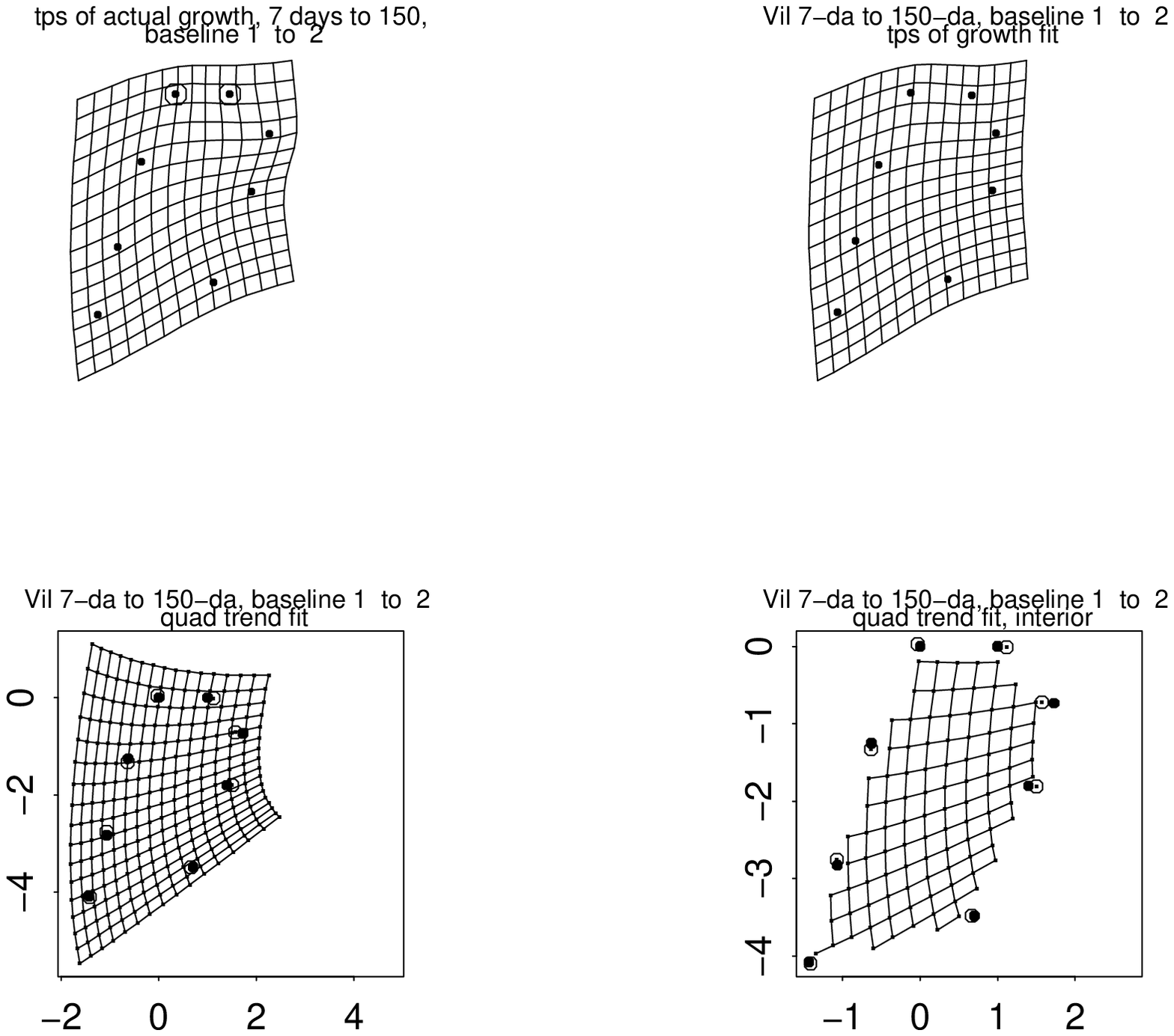}}
\vskip .2in
\noindent{\bf Figure 4.}
\vskip .2in
\centerline{\epsfxsize=7truein\epsffile{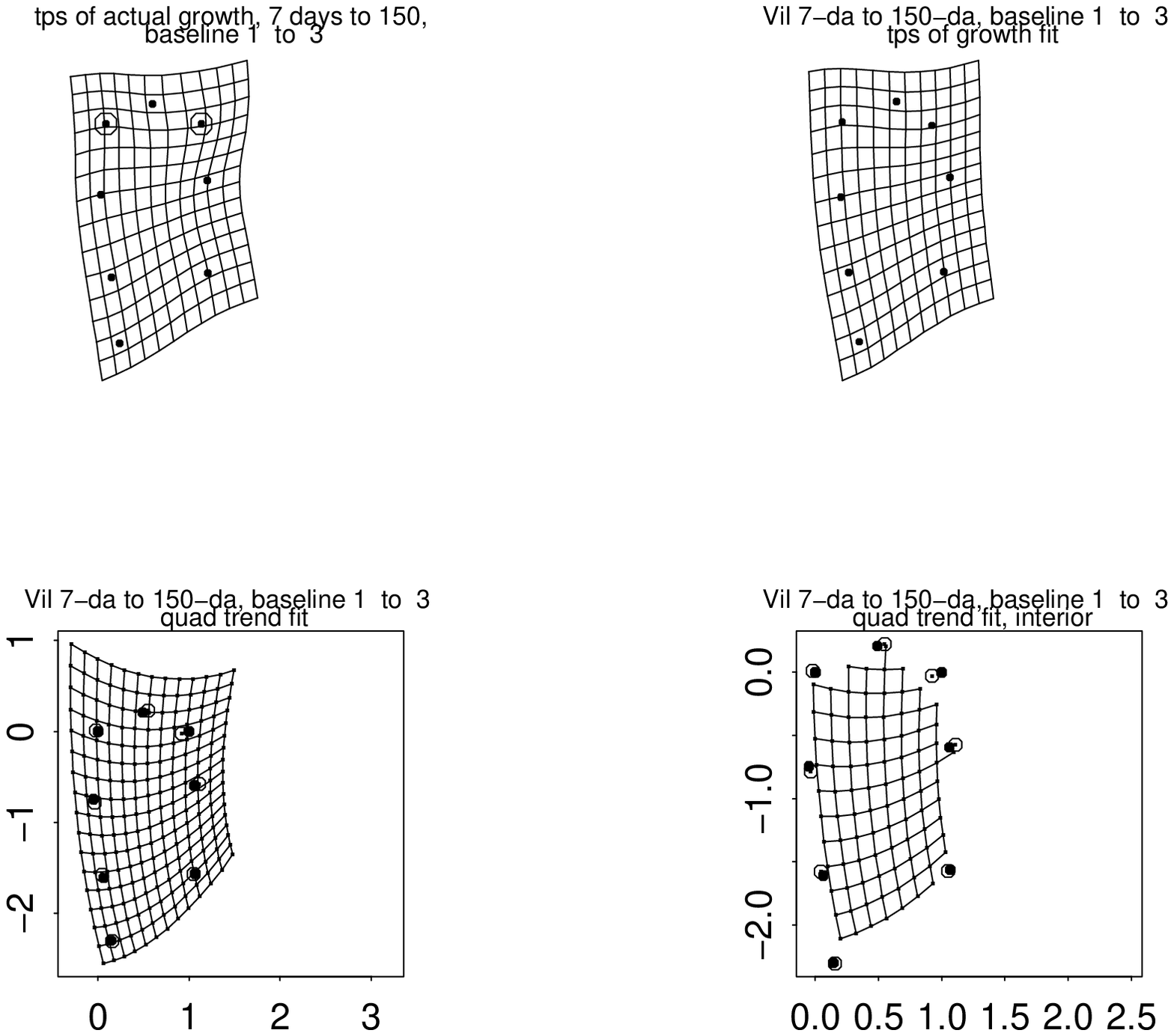}}
\vskip .2in
\noindent{\bf Figure 5.}
\vskip .2in
\centerline{\epsfxsize=7truein\epsffile{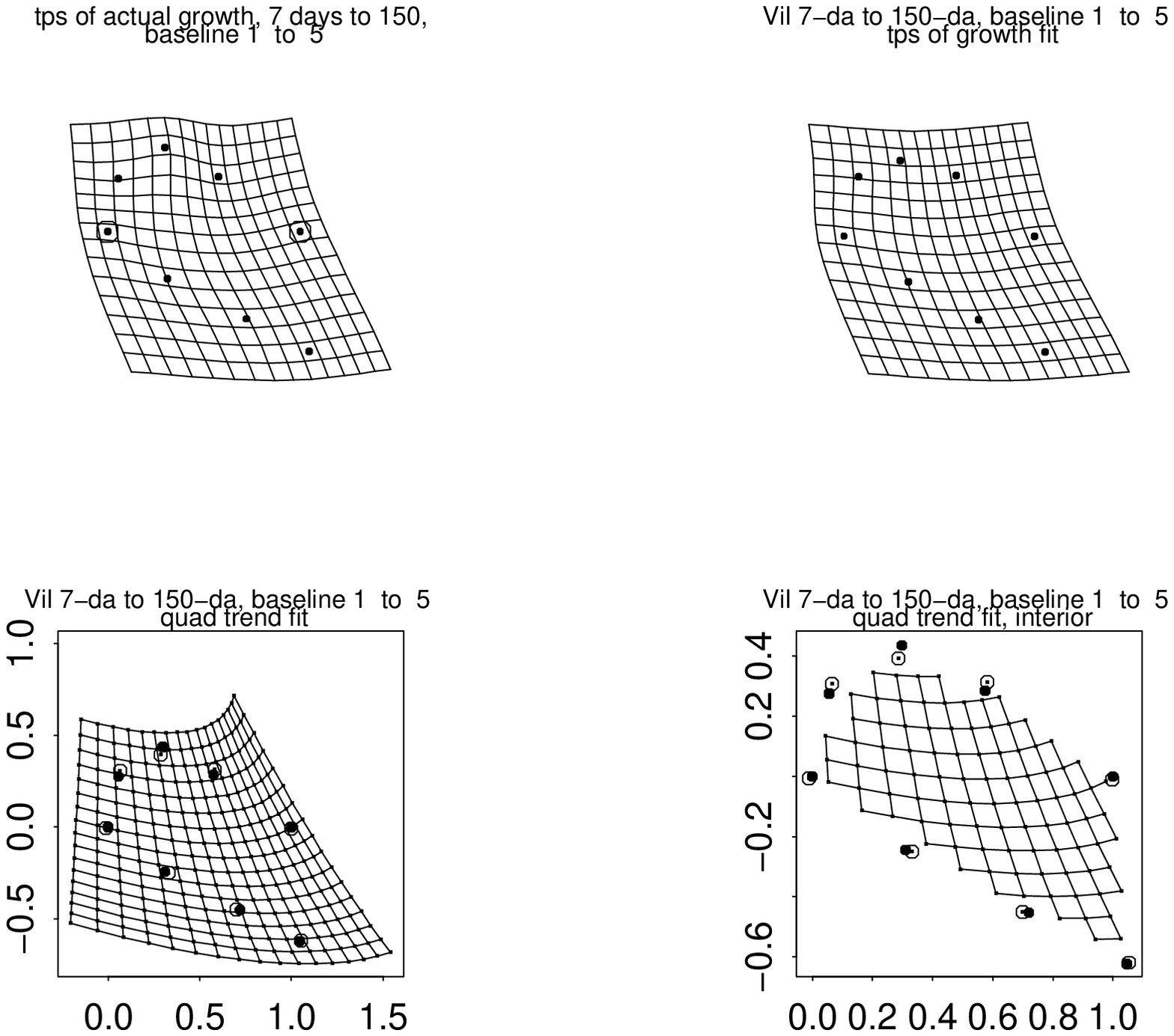}}
\vskip .2in
\noindent{\bf Figure 6.}
\vskip .2in
\centerline{\epsfxsize=7truein\epsffile{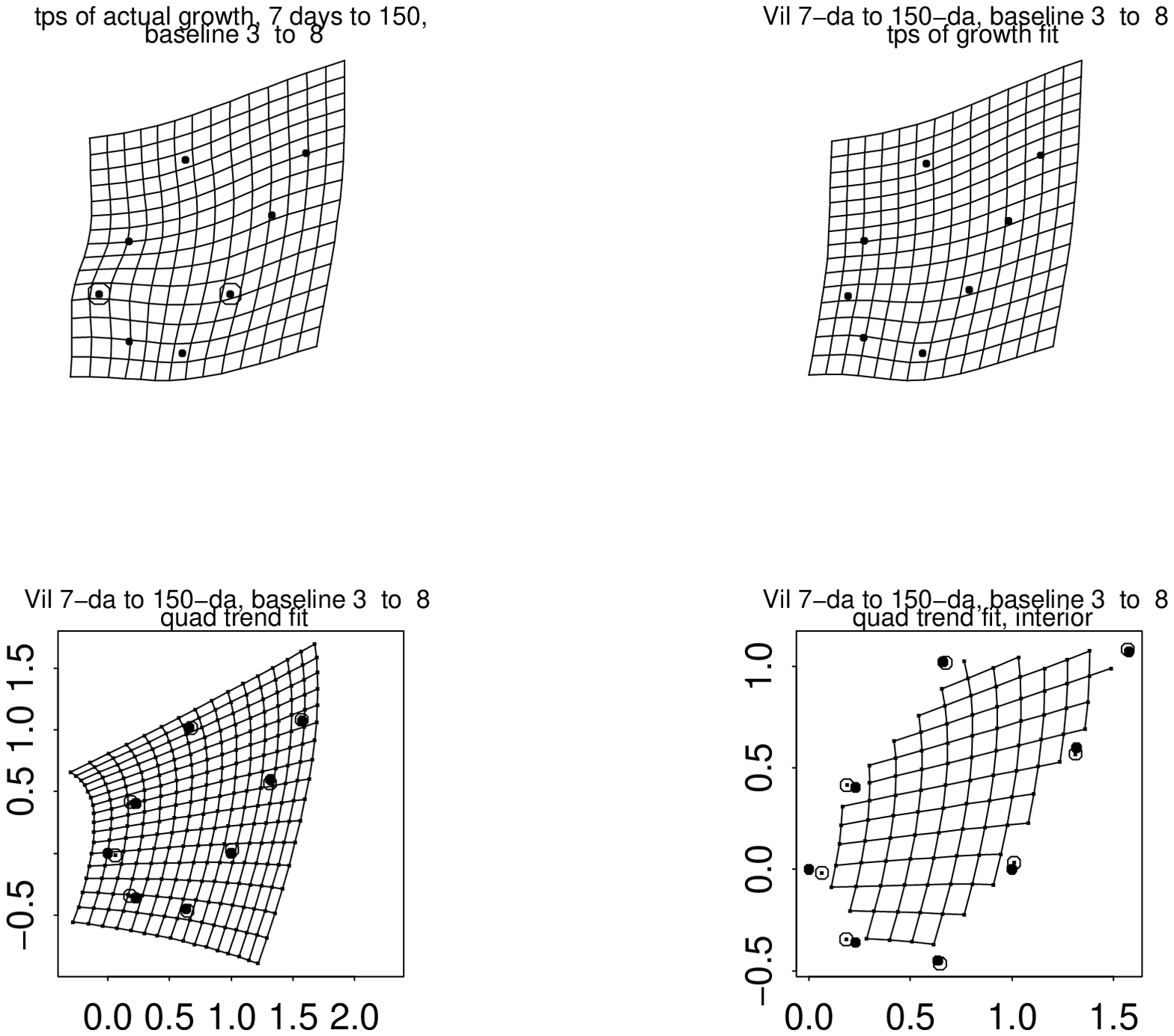}}
\vskip .2in
\noindent{\bf Figure 7.}
\vskip .2in
\centerline{\epsfxsize=7truein\epsffile{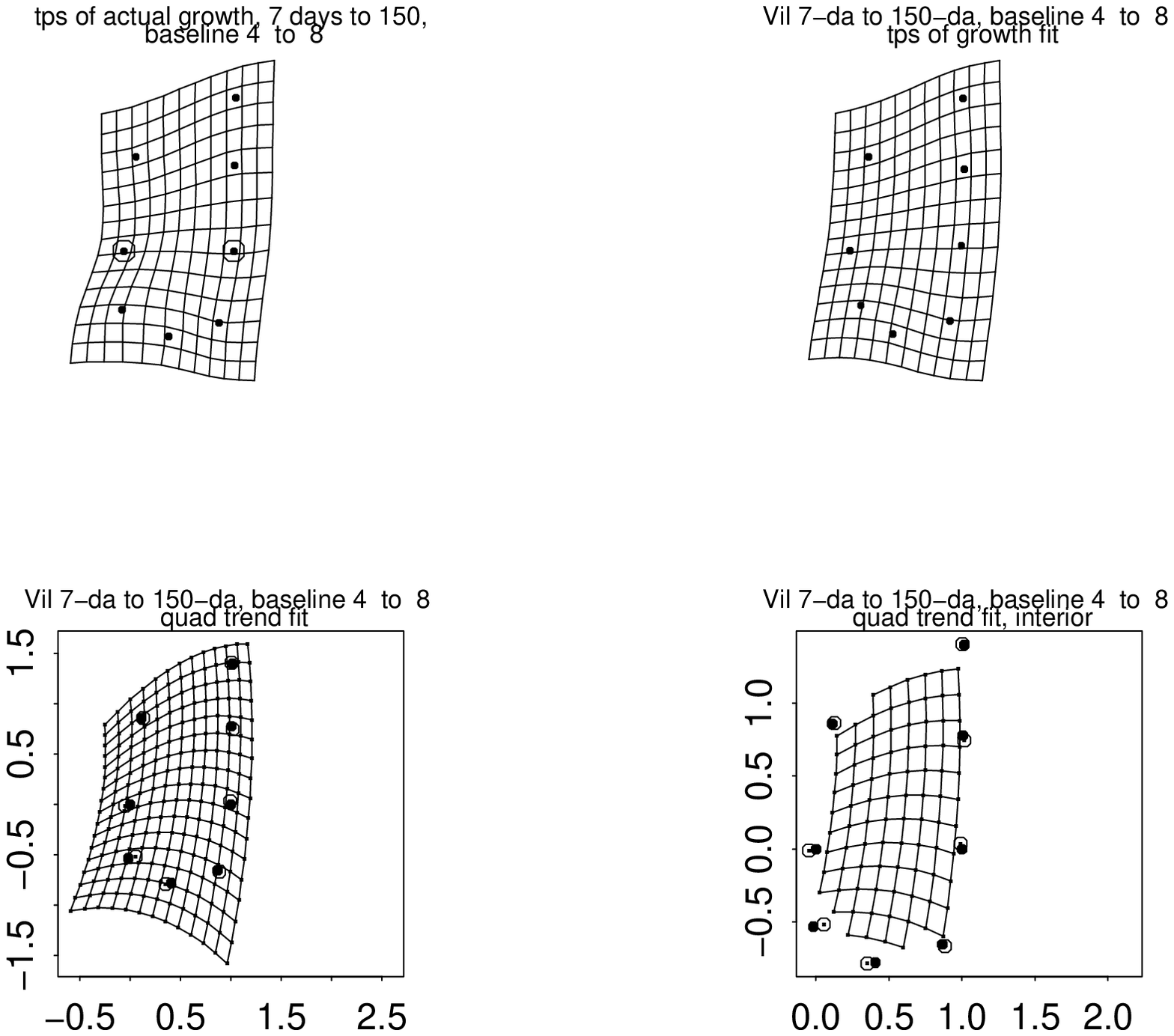}}
\vskip .2in
\noindent{\bf Figure 8.}
\vskip .2in
\centerline{\epsfxsize=7truein\epsffile{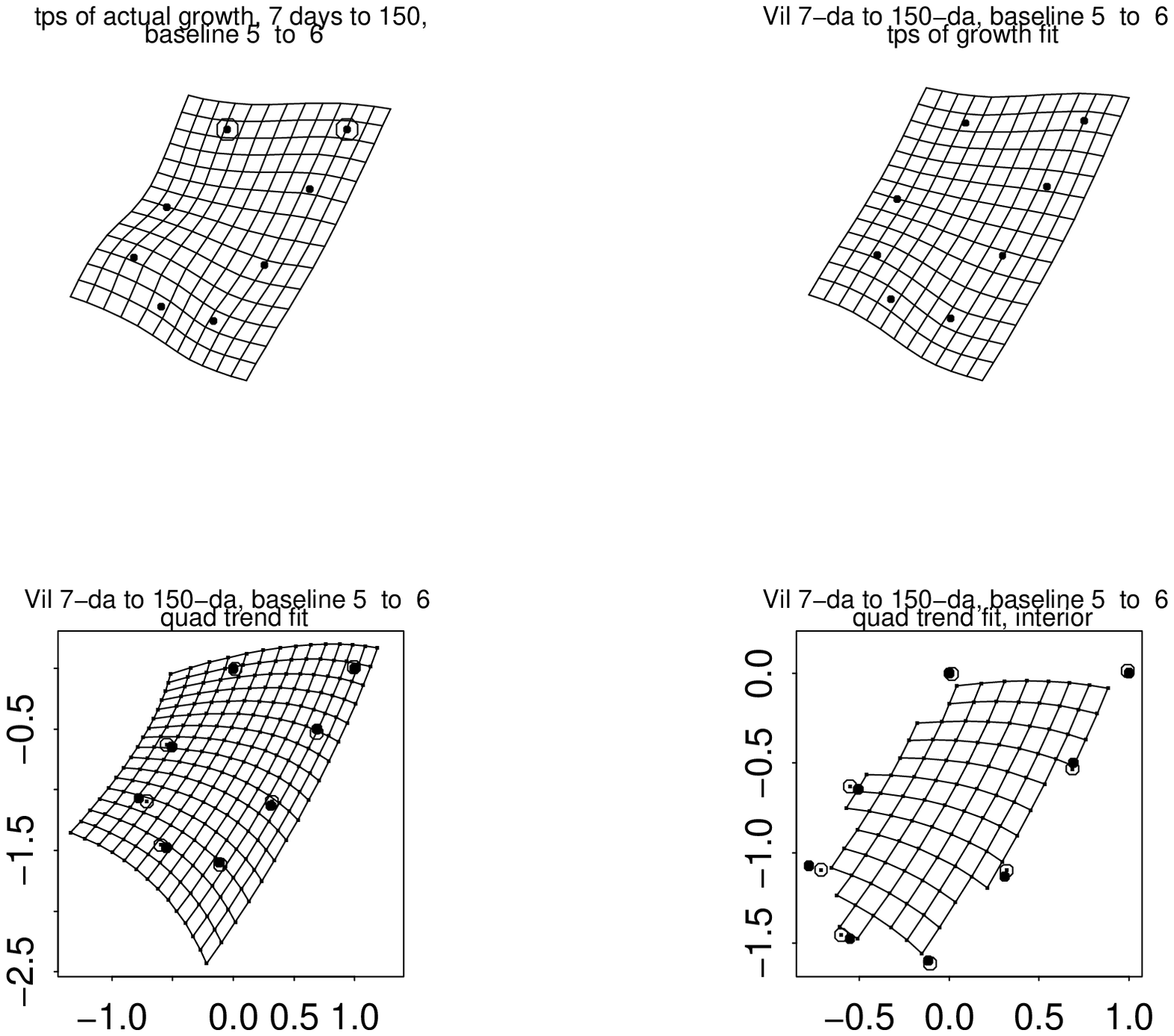}}
\vskip .2in
\noindent{\bf Figure 9.}
\vskip .2in
\centerline{\epsfxsize=7truein\epsffile{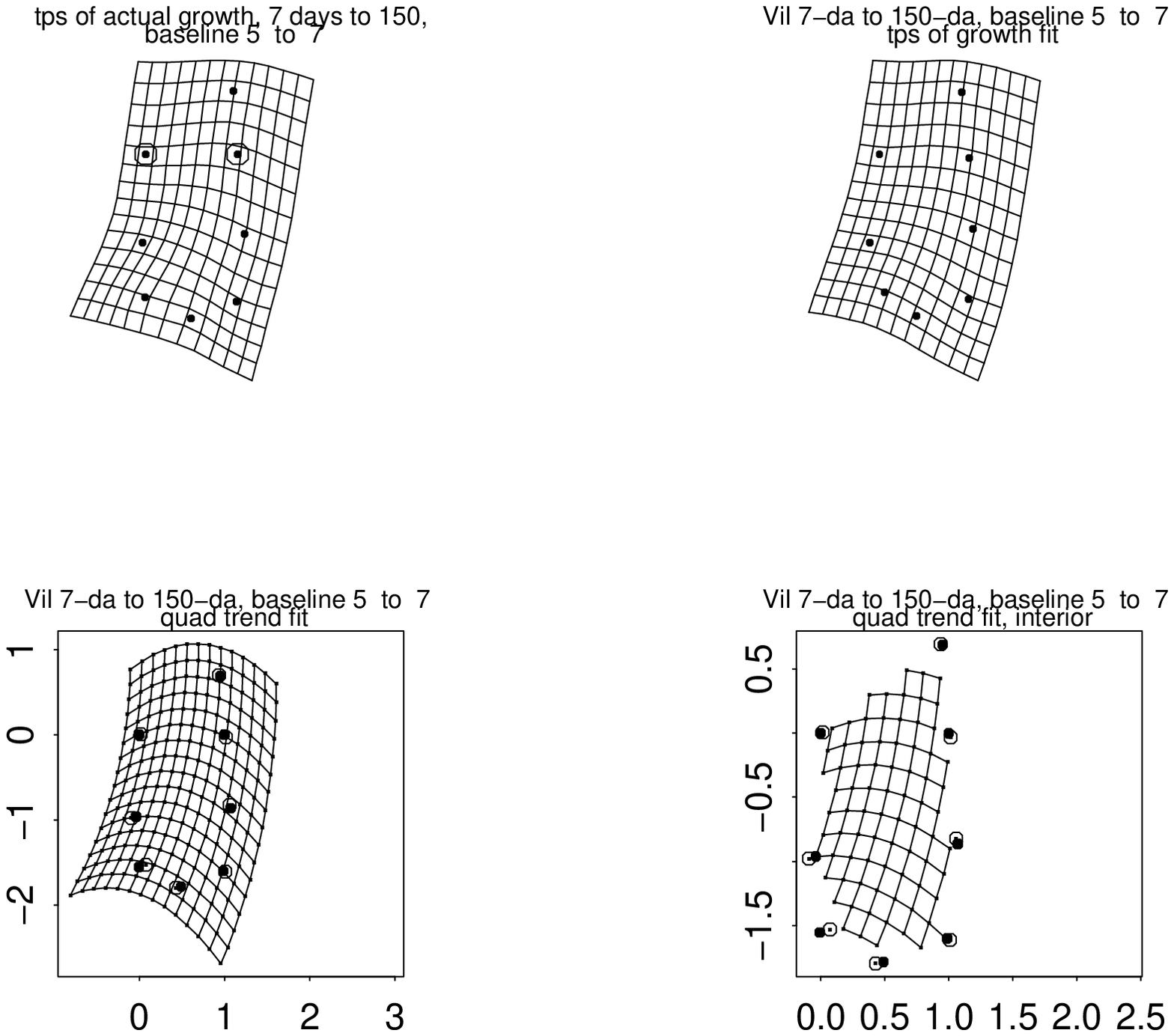}}
\vskip .2in
\noindent{\bf Figure 10.}
\vskip .2in
\centerline{\epsfxsize=7truein\epsffile{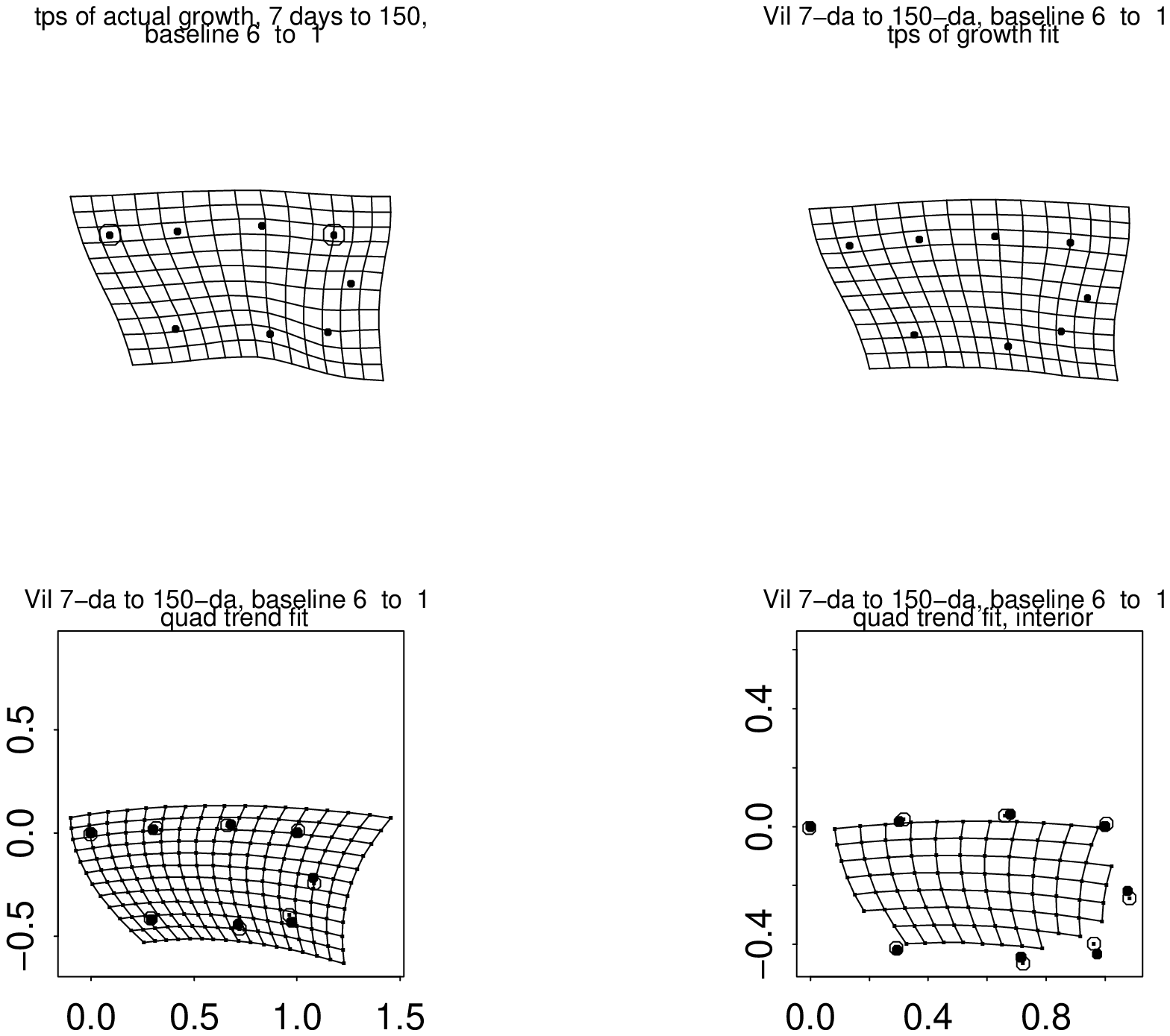}}
\vskip .2in
\noindent{\bf Figure 11.}
\vskip .2in
\centerline{\epsfxsize=6truein\epsffile{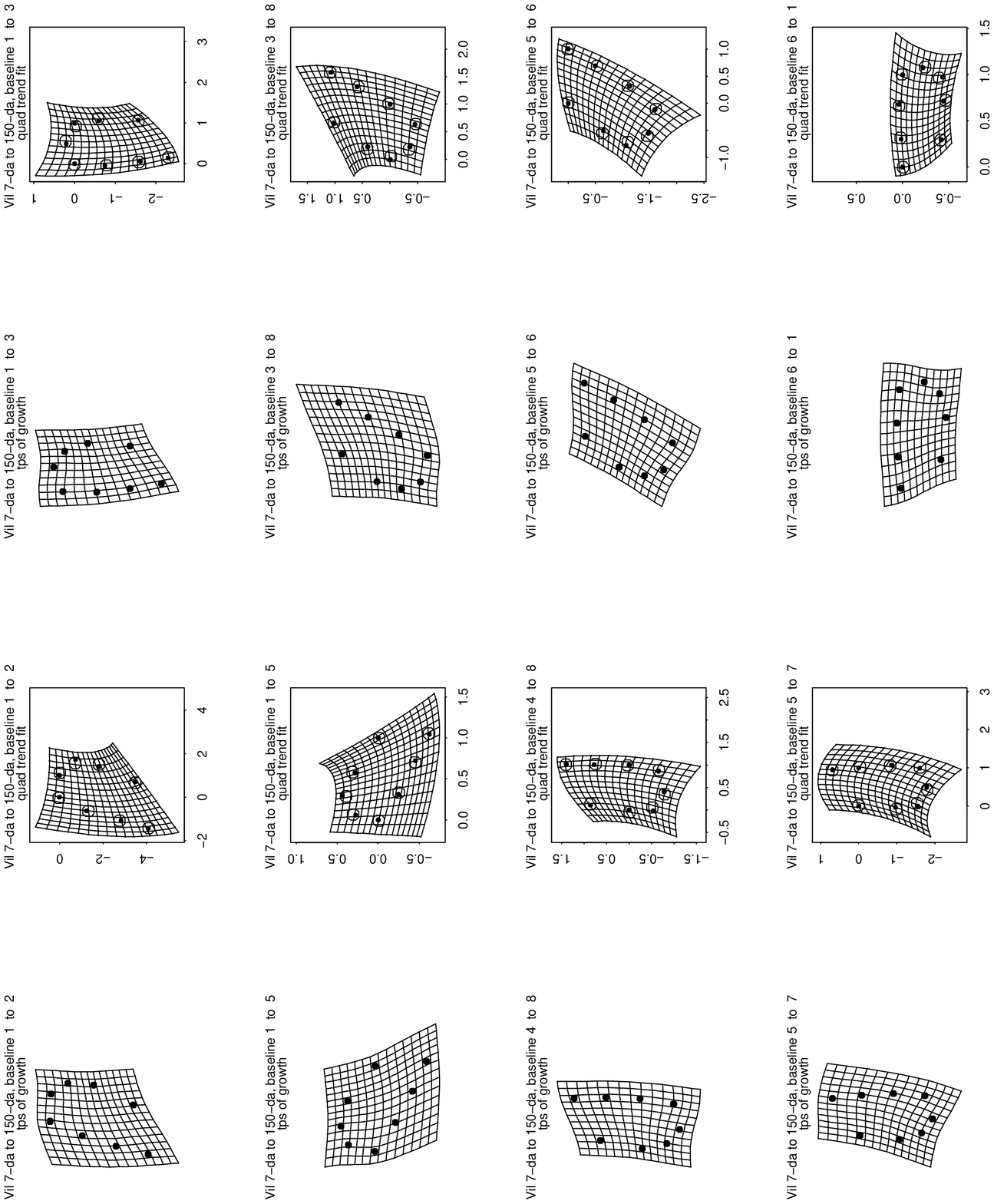}}
\vskip .2in
\noindent{\bf Figure 12.}
\vskip .2in
\centerline{\epsfxsize=6truein\epsffile{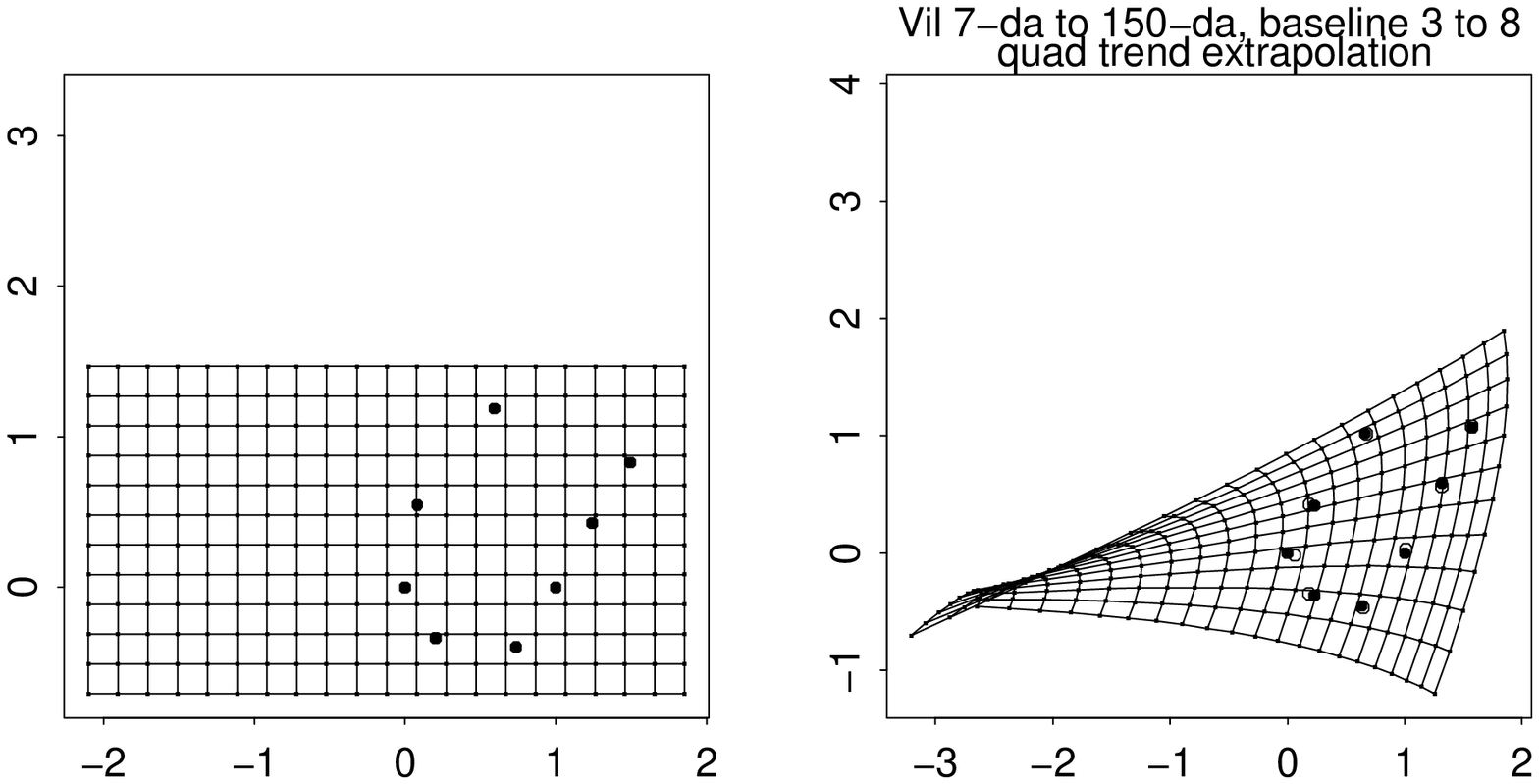}}
\vskip .2in
\noindent{\bf Figure 13.}
\vskip .2in
\centerline{\epsfxsize=6truein\epsffile{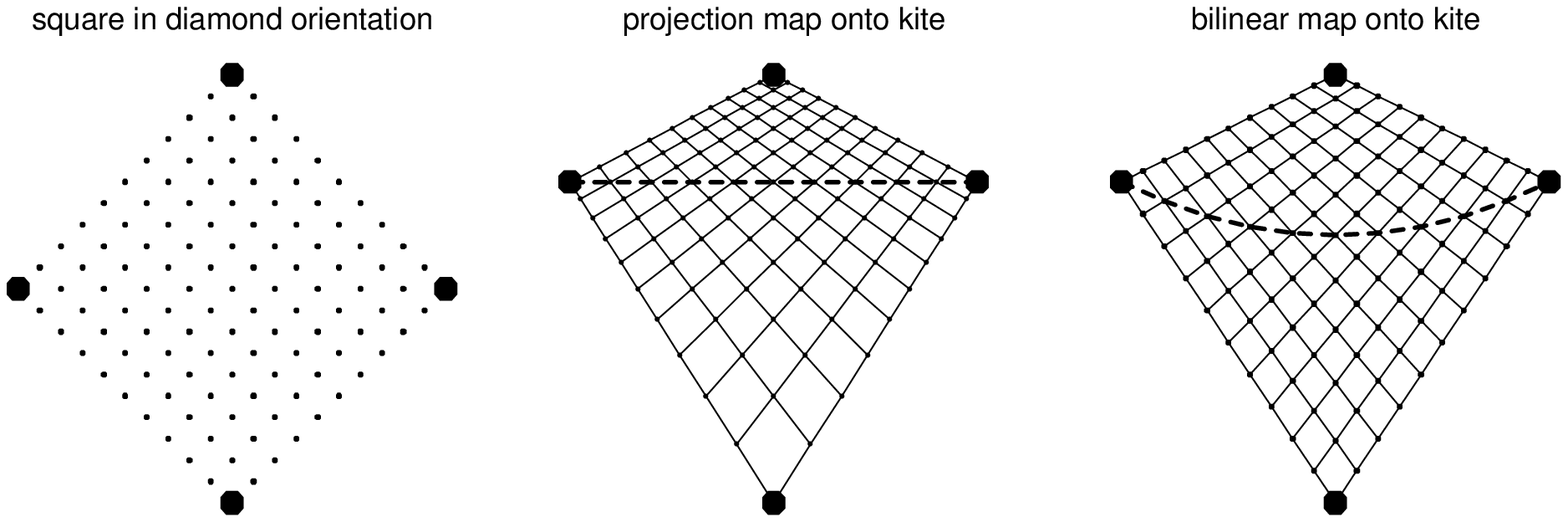}}
\vskip .2in
\noindent{\bf Figure 14.}
\vskip .2in
\centerline{\epsfxsize=6truein\epsffile{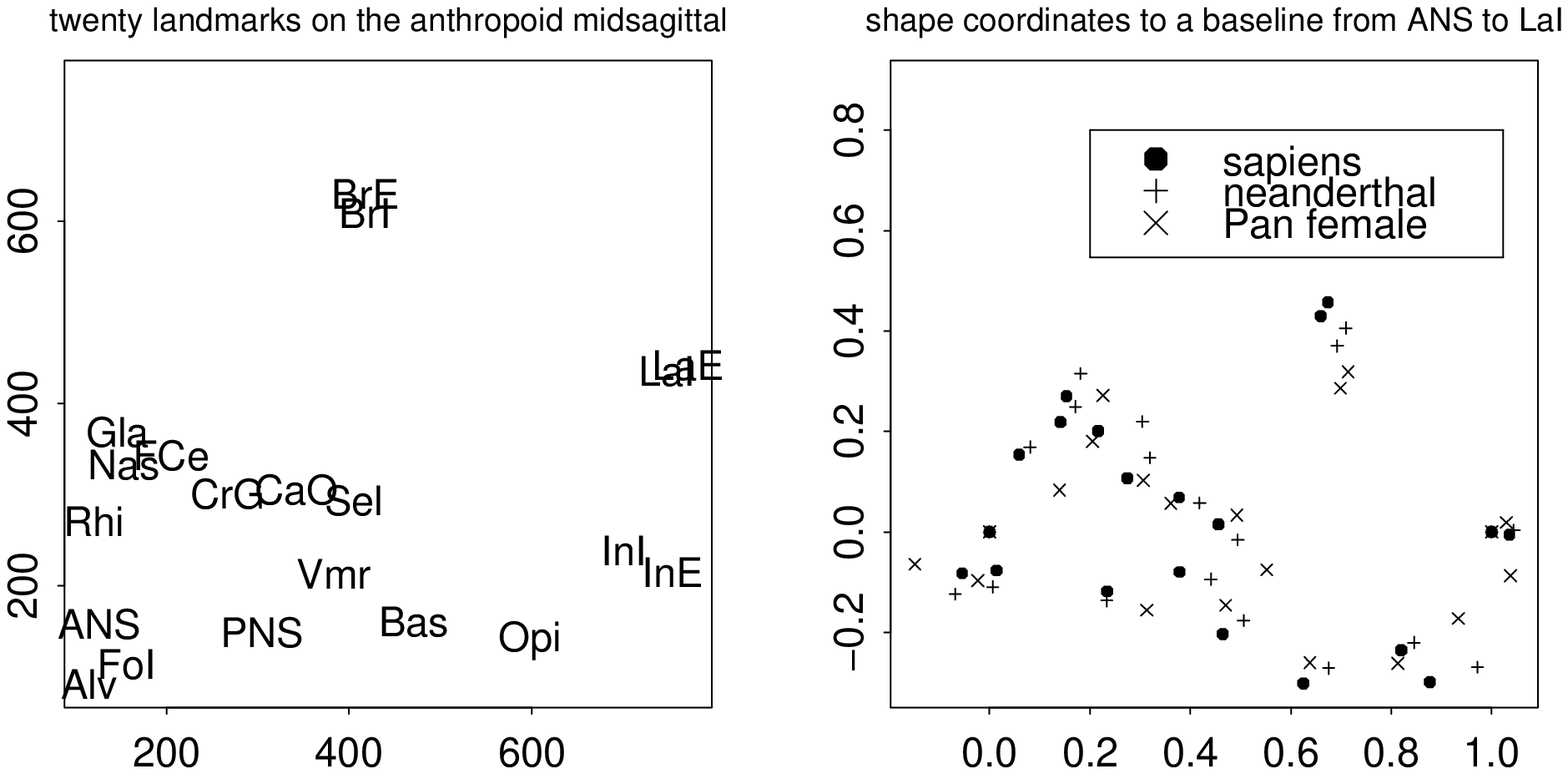}}
\vskip .2in
\noindent {\bf Figure 15.} 
\vskip .2in
\centerline{\epsfxsize=6truein\epsffile{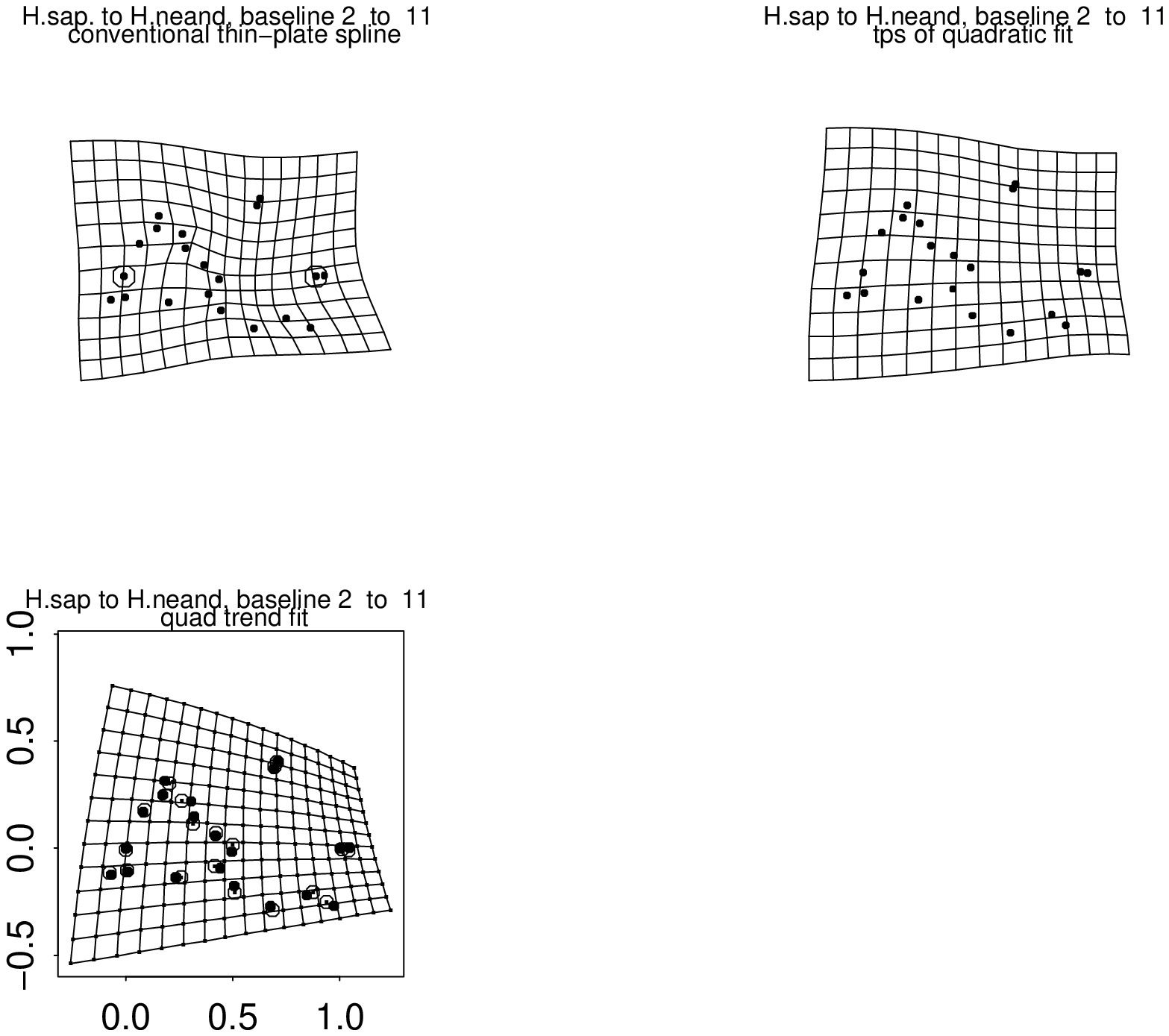}}
\vskip .2in
\noindent{\bf Figure 16.}
\vskip .2in
\centerline{\epsfxsize=6truein\epsffile{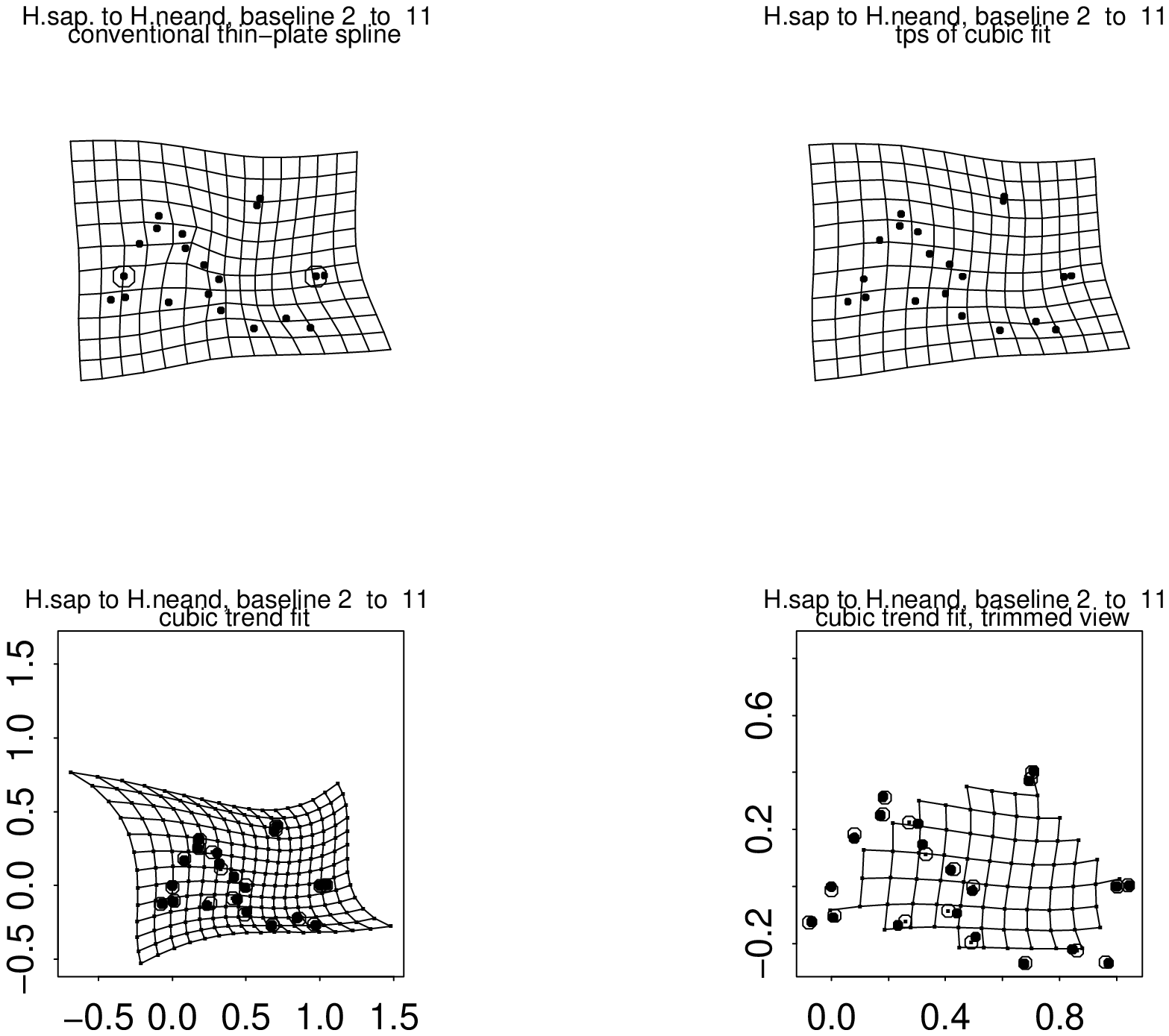}}
\vskip .2in
\noindent{\bf Figure 17.}
\vskip .2in
\centerline{\epsfxsize=6truein\epsffile{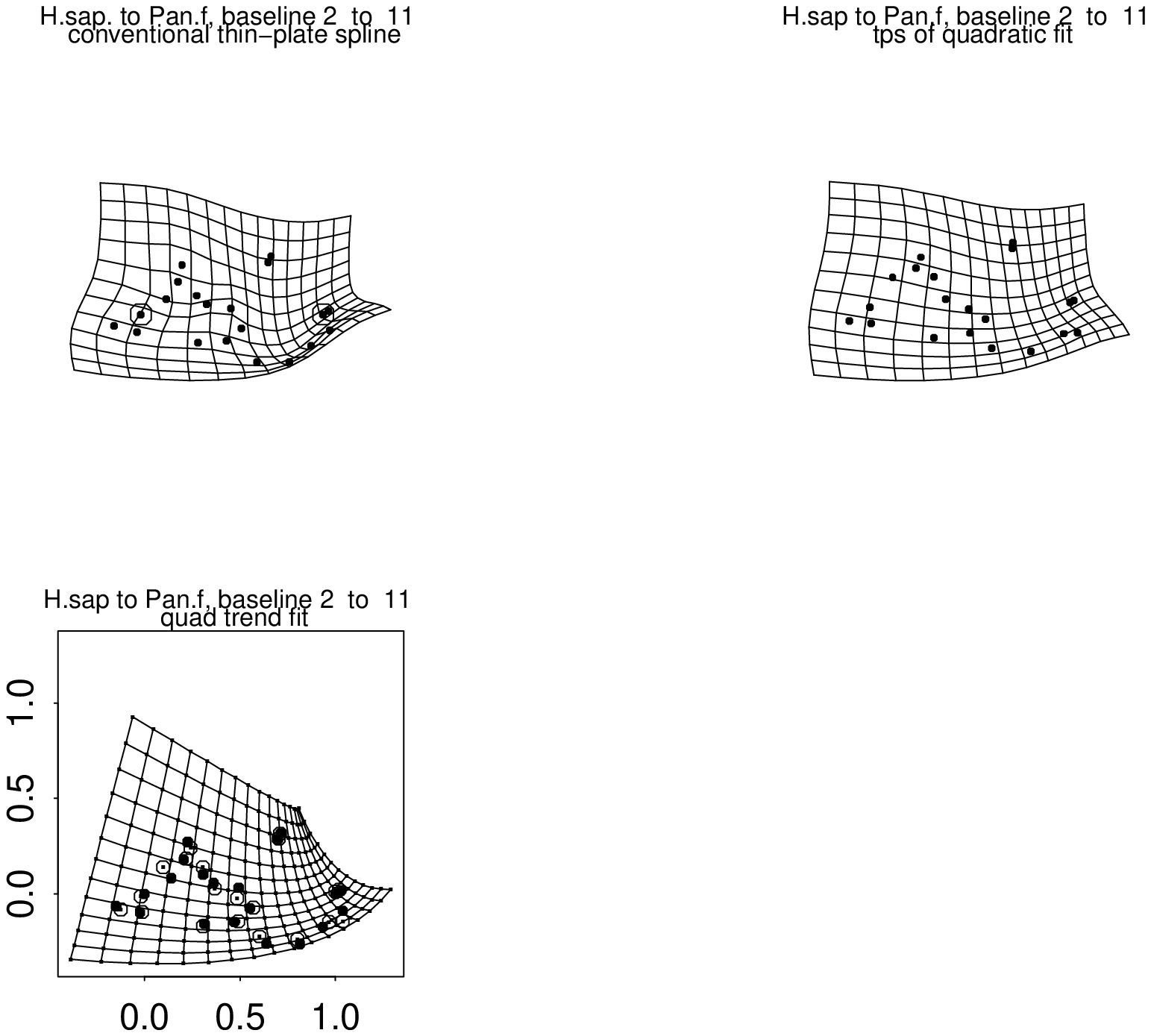}}
\vskip .2in
\noindent{\bf Figure 18.}
\vskip .2in
\centerline{\epsfxsize=6truein\epsffile{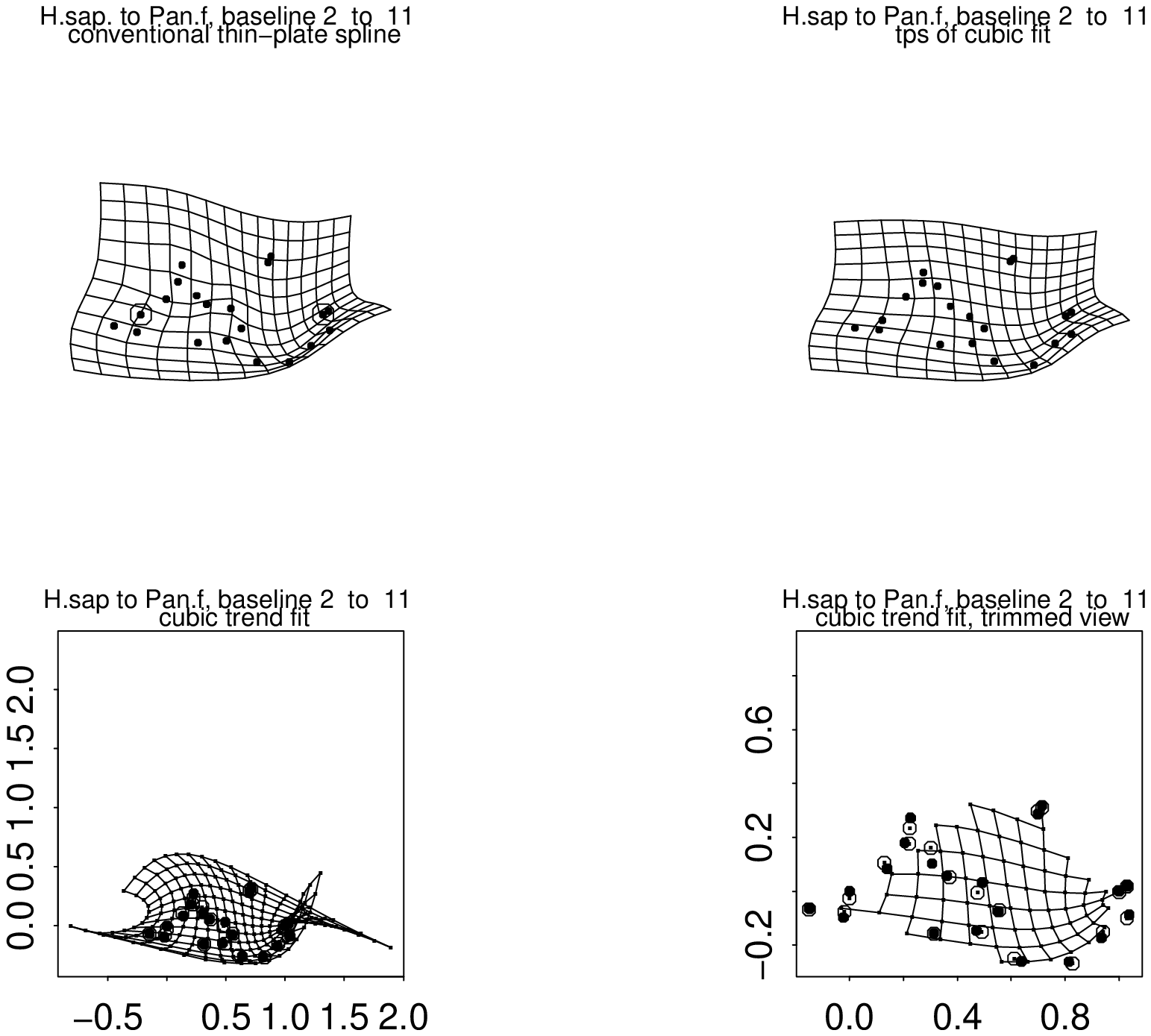}}
\vskip .2in
\noindent{\bf Figure 19.}

\end